\pdfoutput=1

\documentclass[11pt]{article}

\usepackage[final]{acl}

\usepackage{times}
\usepackage{latexsym}

\usepackage[T1]{fontenc}

\usepackage[utf8]{inputenc}

\usepackage{microtype}

\usepackage{inconsolata}

\usepackage{booktabs}
\usepackage{graphicx}
\usepackage{multirow}
\usepackage{makecell}
\usepackage{color}
\usepackage{framed}
\usepackage{amssymb}
\usepackage{amsmath}
\usepackage{colortbl}
\usepackage[utf8]{inputenc}
\usepackage{algorithm}
\usepackage{algorithmic}
\usepackage{subfigure}
\usepackage{subcaption}
\usepackage{caption}
\usepackage{tabularray}

\usepackage{listings}
\definecolor{codegreen}{rgb}{0,0.6,0}
\definecolor{codegray}{rgb}{0.5,0.5,0.5}
\definecolor{codepurple}{rgb}{0.58,0,0.82}
\definecolor{backcolour}{rgb}{0.95,0.95,0.92}

\lstdefinestyle{mystyle}{
    backgroundcolor=\color{backcolour},   
    commentstyle=\color{codegreen},
    keywordstyle=\color{magenta},
    numberstyle=\tiny\color{codegray},
    stringstyle=\color{codepurple},
    basicstyle=\ttfamily\scriptsize,
    breakatwhitespace=false,         
    breaklines=true,                 
    captionpos=b,                    
    keepspaces=true,                 
    numbers=left,                    
    numbersep=5pt,                  
    showspaces=false,                
    showstringspaces=false,
    showtabs=false,                  
    tabsize=2
}

\title{Stealthy Attack on Large Language Model based Recommendation}

\author{
Jinghao Zhang\textsuperscript{1,2}, Yuting Liu\textsuperscript{3}, Qiang Liu\textsuperscript{1,2}, Shu Wu\textsuperscript{1,2}\thanks{To whom correspondence should be addressed.}, Guibing Guo\textsuperscript{3}, Liang Wang\textsuperscript{1,2}, \\
\textsuperscript{1}New Laboratory of Pattern Recognition (NLPR), \\
State Key Laboratory of Multimodal Artificial Intelligence Systems (MAIS), \\
Institute of Automation, Chinese Academy of Sciences \\
\textsuperscript{2}School of Artificial Intelligence, University of Chinese Academy of Sciences \\
  \textsuperscript{3}Northeastern University, China \\
  \texttt{jinghao.zhang@cripac.ia.ac.cn}, \texttt{\{yutingliu\}@stumail.neu.edu.cn} \\ \texttt{\{qiang.liu,shu.wu,wangliang\}@nlpr.ia.ac.cn}, \texttt{\{guogb\}@swc.neu.edu.cn}  \\
  } 

\begin{document}
\newcommand{\themodel}{ATTACK\xspace}
\maketitle
\begin{abstract}
Recently, the powerful large language models (LLMs) have been instrumental in propelling the progress of recommender systems (RS). However, while these systems have flourished, their susceptibility to security threats has been largely overlooked. In this work, we reveal that the introduction of LLMs into recommendation models presents new security vulnerabilities due to their emphasis on the textual content of items. We demonstrate that attackers can significantly boost an item's exposure by merely altering its textual content during the testing phase, without requiring direct interference with the model's training process. Additionally, the attack is notably stealthy, as it does not affect the overall recommendation performance and the modifications to the text are subtle, making it difficult for users and platforms to detect. Our comprehensive experiments across four mainstream LLM-based recommendation models demonstrate the superior efficacy and stealthiness of our approach. Our work unveils a significant security gap in LLM-based recommendation systems and paves the way for future research on protecting these systems. \footnote[1]{The source code will be released at \url{https://github.com/CRIPAC-DIG/RecTextAttack}.}

\end{abstract}

\section{Introduction}
Over the past few decades, recommender systems (RS) have gained considerable significance across various domains. Recently, the powerful large language models (LLMs) have been instrumental in propelling the progress of recommender systems. There has been a notable upswing of interest dedicated to developing LLMs tailored for recommendation task.

Contrary to traditional recommendation models, which rely heavily on abstract and less interpretable ID-based information, LLM-based recommendation models exploit the semantic understanding and strong transferability of LLMs. This approach places a heightened focus on the \textbf{textual content of items}, such as titles and descriptions \cite{lin2023How, chen2023When}. For instance, many researchers \cite{hou2022Universal, hou2023Learning, yuan2023Where, li2023Text, yang2023Collaborative, geng2022Recommendation, cui2022M6Rec, Bao2023TALLRecAE, zhang2023Recommendation, li2023E4SRec, zhang2023CoLLM} have explored modeling user preferences and item characteristics through a linguistic lens. This methodology promises a revolutionary shift in the conventional paradigm of recommendations by providing generalization capabilities to novel items and datasets.

\begin{figure}[t]
    \centering
    \includegraphics[width=\linewidth]{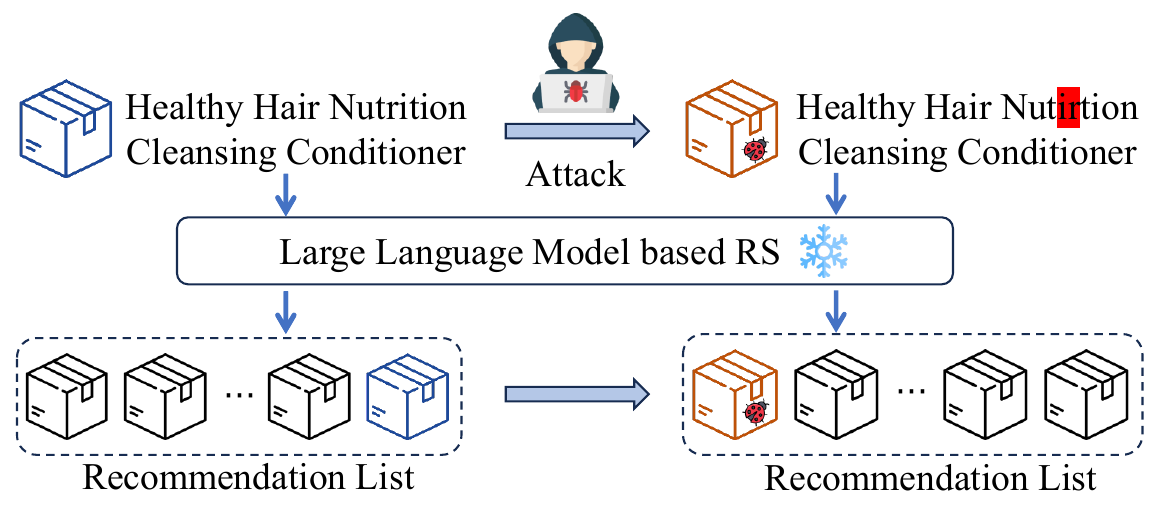}
    \caption{The proposed text attack paradigm on LLM-based RS model. Malicious attackers modify the titles of target items to mislead RS models to rank them higher. The attack is highly stealthy since the modification is subtle and overall recommendation performance is almost unchanged.}
\end{figure}

Despite these advancements, the security of RS remains a largely unaddressed issue. Malicious attacks on these systems can lead to undesirable outcomes, such as the unwarranted promotion of low-quality products in e-commerce platforms or the spread of misinformation in news dissemination contexts. Traditional shilling attack strategies on RS \cite{wang2023RecAD, wang2024Poisoning} involve the generation of fake users who are programmed to give high ratings to specific target items. By introducing such cheating data, it aims at influencing the training of the recommender models and subsequently increasing the exposure of the target items. 

However, the introduction of LLMs into recommendation models presents new security vulnerabilities. In this paper, to the best of our knowledge, we are the first to demonstrate that LLM-based recommendation systems are more vulnerable due to their emphasis on the textual content of items. We demonstrate that\textbf{ attackers can significantly boost an item's exposure by merely altering its textual content during the testing phase}, utilizing simple heuristic re-writing or black-box text attack strategies \cite{morris2020TextAttack}. Compared with traditional shilling attacks, \textbf{this attack paradigm is notably stealthy}, as it does not require influencing the training of the model, and the overall recommendation performance is almost unchanged. Moreover, the modifications to the title are subtle, making it difficult for users and platforms to detect.

We construct comprehensive experiments on four mainstream LLM-based recommendation models \cite{geng2022Recommendation, Bao2023TALLRecAE, li2023Text, zhang2023CoLLM} as victim models to validate the outstanding efficacy and stealthiness of the textual attack paradigm compared with traditional shilling attacks \cite{burke2005Segmentbased,kaur2016Shilling,lin2020Attacking}.  We further delve into the effects of model fine-tuning and item popularity on the attack. Additionally, we investigate the transferability of the attack across various victim models and recommendation tasks to demonstrate its practical applicability and utility in real-world scenarios. Finally, we evaluate a simple re-writing defense strategy, which also can mitigate the issue to some extent.

To summarize our contributions:
\begin{enumerate}
    \item We highlight that LLM-based recommendation models, due to their emphasis on textual content information, could raise previously overlooked security issues.
    \item To the best of our knowledge, we are the first to attack LLM-based recommendation models and  propose the use of textual attacks to promote the exposure of target items.
    \item We perform extensive experiments to demonstrate the efficacy and stealthiness of the textual attack paradigm. Further experiments have revealed the impact of item popularity and model fine-tuning on attacks, as well as explored the transferability of attacks.
    \item Finally, we proposed a simple rewriting defense strategy. While it cannot fully defend against text-based attacks, it can provide some level of defense and contribute to future research.
\end{enumerate}

\section{Method}
In this section, we first introduce the LLM-based recommendation model and formulate the objectives of the attacks. Then, we present two simple model-agnostic text rewriting approaches. Finally, we provide a detailed introduction of black-box text attacks.
\subsection{Problem Definition}
We use the notation $\mathcal{I}= \{i_1, \cdots, i_N\}$ and  $\mathcal{U}=\{u_1, \cdots, u_M\}$ to represent the sets of $N$ items and $M$ users, respectively. Each item $i \in \mathcal{I}$ is associated with textual content $t_i$. Each user $u \in \mathcal{U}$ has interacted with a number of items $\mathcal{I}^u$, indicating that the preference score $y_{ui} = 1$ for $i \in \mathcal{I}^u$.

LLM-based RS models user preference and item feature by transforming user historical behavior sequences $\mathcal{I}^u$ and  target item $i$ into textual prompt $\mathcal{P}_{u,i} = \left[ t_u, t_i, x_u, x_i \right]$, where $t_u = \left[t_{i_1}, \cdots, t_{i_{|\mathcal{I}^u|}}\right]$. $x_u$ and $x_i$ denotes the ID of $i$ and $u$ which are optional in LLM-based RS. We have listed example prompts of four victim models in Table \ref{tab:prompt}. Please refer to Section \ref{sec:victim} for the details of them. The recommendation process can be formulated as: $\hat{y}_{u,i} = f_{\theta}(\mathcal{P}_{u,i})$ where $f_{\theta}$ denotes the LLM-based model.

\begin{table}[]
\resizebox{\linewidth}{!}{%
\begin{tabular}{@{}cc@{}}
\toprule
Model     & Prompt                                                                                                                                                                                                                                                                                                                                                                                                                                                                                                           \\ \midrule
RecFormer & \textless{}HistoryItemTitleList\textgreater{}                                                                                                                                                                                                                                                                                                                                                                                                                                                                           \\ \midrule
P5        & \begin{tabular}[c]{@{}c@{}}I would like to recommend some  items for   \textless{}UserID\textgreater{}. Is the following \\ item a good choice?  \{TargetItemTitle\}\end{tabular}                                                                                                                                                                                                                                                                                                                                \\ \midrule
TALLRec   & \makecell[c]{A user has given high ratings to the following products: <HistoryItemTitleList>. \\ Leverage the information to predict whether the user would enjoy \\ the product titled <TargetItemTitle>? Answer with "Yes" or "No".}                                                                                                                                                                                                   \\ \midrule
CoLLM     & \makecell[c]{A user has given high ratings to the following products: <HistoryItemTitleList>. \\ Additionally, we have information about the user's preferences encoded \\ in the feature <UserID>. Using all available information, make a prediction \\ about whether the user would enjoy the product titled <TargetItemTitle> \\ with the feature <TargetItemID>? Answer with "Yes" or "No"} \\ \bottomrule
\end{tabular}}
\caption{Prompts $\mathcal{P}_{u,i}$ of four victim models. P5 unifies different recommendation tasks with different prompts and we only show one example.}
\label{tab:prompt}
\end{table}

The goal of the attack task is to promote target items $\mathcal{I}'$ (increasing the exposure or user interaction probability) through imperceptibly modifying their textual content (we use title in this work).

\subsection{Victim Model-Agnostic Attack}
In this subsection, we first introduce two simple, victim model-agnostic strategies employed for altering item textual content to make them more linguistically attractive to users. Our approaches include trivial attack with word insertion and re-writing leveraging Generative Pre-trained Transformers (GPTs).

\subsubsection{Trivial Attack with Word Insertion}
The core premise of this strategy is founded on the assumption that positive or exclamatory words can attract users. By infusing item titles with a select number of positive words, we aim to increase the items' attractiveness and, consequently, their likelihood of being recommended by the system. Specifically, we randomly select $k$ words form a pre-defined word corpus which is common-used in item titles. These selected words are then inserted to the end of the original text content to retain the overall coherence.

\begin{center}
    \fcolorbox{black}{gray!20}{ \parbox{0.9\linewidth}{
    \textbf{Positive word corpus}: [`good', `great', `best', `nice', `excellent', `amazing', `awesome', `fantastic', `wonderful', `perfect', `ultimate', `love', `like', `beautiful', `well', `better', `easy', `happy', `recommend', `works', `fine', `fast', `fun', `price', `quality', `product', `value', `bought', `purchase', `top', `popular', `choice', `!!!' ]
    }}
    \label{c:p}
\end{center}

\subsubsection{Re-writing with GPTs}
While the insertion of positive words offers a straightforward means of enhancing content appeal, it can sometimes result in awkward or forced phrasings that diminish the content's natural flow and potentially arouse user suspicion. To address these shortcomings, we propose to use GPTs to rewrite the content of items in a more attractive way by leveraging its rich common sense knowledge and powerful generation capabilities. Specifically, we instruct GPT-3.5-turbo with the following prompts to generate attractive and fluency titles.

\begin{center}
    \fcolorbox{black}{gray!20}{ \parbox{0.9\linewidth}{
    \textbf{Prompt 1}: You are a marketing expert that helps to promote the product selling. Rewrite the product title in <MaxLen> words to keep its body the same but more attractive to customers: <ItemTitle>. \\
    \textbf{Prompt 2}: Here is a basic title of a product. Use your creativity to transform it into a catchy and unique title in <MaxLen> words that could attract more attention: <ItemTitle>. \\
    \textbf{Prompt 3}: Rewrite this product's title by integrating positive and appealing words, making it more attractive to potential users without altering its original meaning (in <MaxLen> words): <ItemTitle>.
    }}
\end{center}

\subsection{Exploring Vulnerabilities in LLM-Based Recommendation Models through Black-Box Text Attacks}
In this subsection, we present an examination of traditional black-box text attack methods to explore the vulnerabilities within LLM-based recommendation models. Black-box text attack methods typically involve manipulating or perturbing text inputs to deceive or mislead a natural language processing (NLP) model while having no access to the model's internal parameters or gradients. The goal of such attacks is mathematically formulated as:
\begin{equation}
    \label{eq:goal}
    \underset{t_{i}'}{\arg \max } \, \mathbb{E}_{u \in \mathcal{U}'} \, f_{\theta}(\mathcal{P}_{u,i}'), 
\end{equation}
where $\mathcal{P}_{u,i}^{'} = \left[ t_u, t_i', x_u, x_i \right]$ denotes the prompts consisting of the user text $t_u$ and the manipulated title of the target item $t_i'$. Following the framework proposed by \citet{morris2020TextAttack}, text attacks are comprised of four principal components:
\begin{itemize}
    \item Goal Function: This function evaluates the effectiveness of the perturbed input $x'$ in achieving a specified objective, serving as a heuristic for the search method to identify the optimal solution. In this study, the aim is to promote the target items as in Equation \ref{eq:goal}.
    \item Constraints: These are conditions that ensure the perturbations remain valid alterations of the original input, emphasizing aspects such as semantic retention and maintaining consistency in part-of-speech tags.
    \item Transformation: A process that applies to an input to generate possible perturbations, which could involve strategies like swapping words with similar ones based on word embeddings, using synonyms from a thesaurus, or substituting characters with homoglyphs.
    \item Search Method: This method involves iteratively querying the model to select promising perturbations generated through transformations, employing techniques such as a greedy approach with word importance ranking, beam search, or a genetic algorithm.
\end{itemize}
While the specific components of text attack methodologies may vary, the overarching framework remains consistent, as depicted in Algorithm 1. In this work, we have implemented four widely-used attacks: DeepwordBug \cite{gao2018black}, TextFooler \cite{jin2020bert}, BertAttack \cite{li2020bert}, and PuncAttack \cite{formento2023using}. DeepwordBug and PuncAttack are character-level which manipulate texts by introducing typos and inserting punctuation. TextFooler and BertAttack are word-level that aim to replace words with synonyms or contextually similar words. Please refer to the appendix \ref{app:attack} for the details of text attack paradigm and these four methods.

\begin{algorithm}
\caption{Text Attack Framework}
\label{alg:attack}
\begin{algorithmic}[1]
\REQUIRE Original text $x$, Target model $M$, Goal function $G$, Constraints $C$, Transformations $T$, Search Method $S$
\ENSURE Adversarial text $x'$, Adversarial score $G.score(x')$

\STATE Initialize $x'$ as a copy of $x$.
\WHILE{not $S$.StoppingCriteria()}
    \STATE Select a transformation $t$ from allowable transformations $T$.
    \STATE Generate $x'$ by applying $t$ to $x$.
    \IF{$C$.Satisfied($x'$)}
        \IF{$S$.AchieveGoal($x'$)}
            \STATE \textbf{return} $x'$, $G$.score($x'$).
        \ENDIF
    \ENDIF
\ENDWHILE
\end{algorithmic}
\end{algorithm}

\section{Experiments}
\begin{table*}[]
\resizebox*{\textwidth}{!}{
\begin{tabular}{@{}cccccccccc@{}}
\toprule
\multicolumn{1}{l}{\multirow{2}{*}{Dataset}} & \multirow{2}{*}{Method}          & \multicolumn{3}{c}{Effectiveness}                                              & \multicolumn{5}{c}{Stealthiness}                                                                                    \\ \cmidrule(l){3-5}  \cmidrule(l){6-10} 
\multicolumn{1}{l}{}                         &                                  & Exposure $\uparrow$ & Rel. Impro. $\uparrow$ & \# queries $\downarrow $           & NDCG@10 $\uparrow$ & Cos. $\uparrow$ & Rouge-l $\uparrow$ & Perplexity $\downarrow $ & \# pert. words $\downarrow $ \\ \midrule
\multicolumn{1}{c|}{\multirow{7}{*}{Sports}} & \multicolumn{1}{c|}{Clean}       & 0.00282             & -               & \multicolumn{1}{c|}{-}            & 0.00780            & 1.000           & 1.000              & 2158.7                   & -                          \\
\multicolumn{1}{c|}{}                        & \multicolumn{1}{c|}{ChatGPT}     & 0.00293             & 3.9\%             & \multicolumn{1}{c|}{-}            & 0.00781            & 0.794           & 0.499              & \textbf{1770.9}          & -                           \\
\multicolumn{1}{c|}{}                        & \multicolumn{1}{c|}{Trivial}      & 0.00242             & -14.4\%             & \multicolumn{1}{c|}{-}            & \textbf{0.00782}   & \textbf{0.896}  & \textbf{0.869}     & 4376.6                   & -                           \\
\multicolumn{1}{c|}{}                        & \multicolumn{1}{c|}{Deepwordbug} & 0.01488             & 427.2\%             & \multicolumn{1}{c|}{\textbf{38.6}} & 0.00757            & 0.702           & 0.451              & 5595.1                   & 4.3                          \\
\multicolumn{1}{c|}{}                        & \multicolumn{1}{c|}{TextFooler}  & \textbf{0.01547}    & \textbf{448.2\%}    & \multicolumn{1}{c|}{87.5}          & 0.00780            & 0.758           & 0.575              & 2070.8                   & 3.4                          \\
\multicolumn{1}{c|}{}                        & \multicolumn{1}{c|}{PuncAttack}  & 0.01138             & 303.3\%             & \multicolumn{1}{c|}{52.6}          & \textbf{0.00762}   & 0.857           & 0.635              & 2410.9                   & \textbf{2.9}                 \\
\multicolumn{1}{c|}{}                        & \multicolumn{1}{c|}{BertAttack}  & 0.01371             & 385.8\%             & \multicolumn{1}{c|}{141.7}         & 0.00781            & 0.850           & 0.679              & 7760.1                   & \textbf{2.8}                 \\ \midrule
\multicolumn{1}{c|}{\multirow{7}{*}{Beauty}} & \multicolumn{1}{c|}{Clean}       & 0.00458             & -               & \multicolumn{1}{c|}{-}            & 0.01258            & 1.000           & 1.000              & 611.6                    & -                          \\
\multicolumn{1}{c|}{}                        & \multicolumn{1}{c|}{ChatGPT}     & 0.00583             & 27.2\%              & \multicolumn{1}{c|}{-}            & 0.01197            & 0.822           & 0.516              & \textbf{501.8}           & -                           \\
\multicolumn{1}{c|}{}                        & \multicolumn{1}{c|}{Trivial}      & 0.00389             & -15.2\%             & \multicolumn{1}{c|}{-}            & \textbf{0.01249}   & \textbf{0.939}  & \textbf{0.901}     & 1189.5                   & -                           \\
\multicolumn{1}{c|}{}                        & \multicolumn{1}{c|}{Deepwordbug} & 0.02134             & 365.4\%             & \multicolumn{1}{c|}{\textbf{47.0}} & \textbf{0.01257}   & 0.806           & 0.649              & 2261.5                   & 3.7                          \\
\multicolumn{1}{c|}{}                        & \multicolumn{1}{c|}{TextFooler}  & \textbf{0.02844}    & \textbf{520.4\%}    & \multicolumn{1}{c|}{104.1}         & 0.01224            & 0.816           & 0.640              & 960.9                    & 3.7                          \\
\multicolumn{1}{c|}{}                        & \multicolumn{1}{c|}{PuncAttack}  & 0.01654             & 260.8\%             & \multicolumn{1}{c|}{72.7}          & \textbf{0.01257}   & 0.881           & 0.734              & 1018.3                   & \textbf{2.8}                 \\
\multicolumn{1}{c|}{}                        & \multicolumn{1}{c|}{BertAttack}  & \textbf{0.02705}    & \textbf{490.0\%}    & \multicolumn{1}{c|}{208.8}         & 0.01213            & 0.863           & 0.709              & 1764.4                   & 3.2                          \\ \midrule
\multicolumn{1}{c|}{\multirow{7}{*}{Toys}}   & \multicolumn{1}{c|}{Clean}       & 0.00439             & -               & \multicolumn{1}{c|}{-}            & 0.02380            & 1.000           & 1.000              & 4060.4                   & -                          \\
\multicolumn{1}{c|}{}                        & \multicolumn{1}{c|}{ChatGPT}     & 0.00547             & 24.7\%              & \multicolumn{1}{c|}{-}            & \textbf{0.02369}   & 0.793           & 0.454              & \textbf{1967.1}          & -                           \\
\multicolumn{1}{c|}{}                        & \multicolumn{1}{c|}{Trivial}      & 0.00439             & 0.0\%               & \multicolumn{1}{c|}{-}            & 0.02366            & \textbf{0.880}  & \textbf{0.852}     & 7874.4                   & -                           \\
\multicolumn{1}{c|}{}                        & \multicolumn{1}{c|}{Deepwordbug} & 0.01595             & 263.5\%             & \multicolumn{1}{c|}{\textbf{33.4}} & 0.02365            & 0.697           & 0.511              & 8581.1                   & 3.3                          \\
\multicolumn{1}{c|}{}                        & \multicolumn{1}{c|}{TextFooler}  & \textbf{0.02232}    & \textbf{408.8\%}    & \multicolumn{1}{c|}{84.1}          & 0.02366            & 0.714           & 0.510              & 3702.0                   & 3.4                          \\
\multicolumn{1}{c|}{}                        & \multicolumn{1}{c|}{PuncAttack}  & 0.01241             & 182.9\%             & \multicolumn{1}{c|}{41.3}          & \textbf{0.02328}   & 0.862           & 0.675              & 3747.5                   & \textbf{2.2}                 \\
\multicolumn{1}{c|}{}                        & \multicolumn{1}{c|}{BertAttack}  & 0.01384             & 215.6\%             & \multicolumn{1}{c|}{120.3}         & 0.02340            & 0.854           & 0.693              & 10363.0                  & \textbf{2.4}                 \\ \bottomrule
\end{tabular}
}
\caption{Performance comparison of attacking \textbf{Recformer} where Rel. Impro. denotes relative improvement against clean setting. The best result is in \textbf{boldface}.}
\label{tab:recformer}
\end{table*}
\subsection{Experimental Settings}
\subsubsection{Victim Models}
\label{sec:victim}
We choose four mainstream LLM-based recommendation models as our victim models: \textbf{Recformer} \cite{li2023Text}, \textbf{P5} \cite{geng2022Recommendation}, \textbf{TALLRec} \cite{Bao2023TALLRecAE} and \textbf{CoLLM} \cite{zhang2023CoLLM}. Please refer to Appendix \ref{app:victim} for more details.

\subsubsection{Compared Shilling Attacks}
Shilling attacks aim to generate fake users that assign high ratings for a target item, while also rating other items to act like normal users for evading. We compare our text attack paradigm with white-box shilling attacks \textbf{Random attack} \cite{kaur2016Shilling}, \textbf{Bandwagon attack} \cite{burke2005Segmentbased}, and gray-box \textbf{Aush} \cite{lin2020Attacking} and \textbf{Leg-UP} \cite{lin2024ShillingBlackbox}. Please refer to Appendix \ref{app:baselines} for more details.

\begin{figure*}[t]
    \centering
    \includegraphics[width=\textwidth]{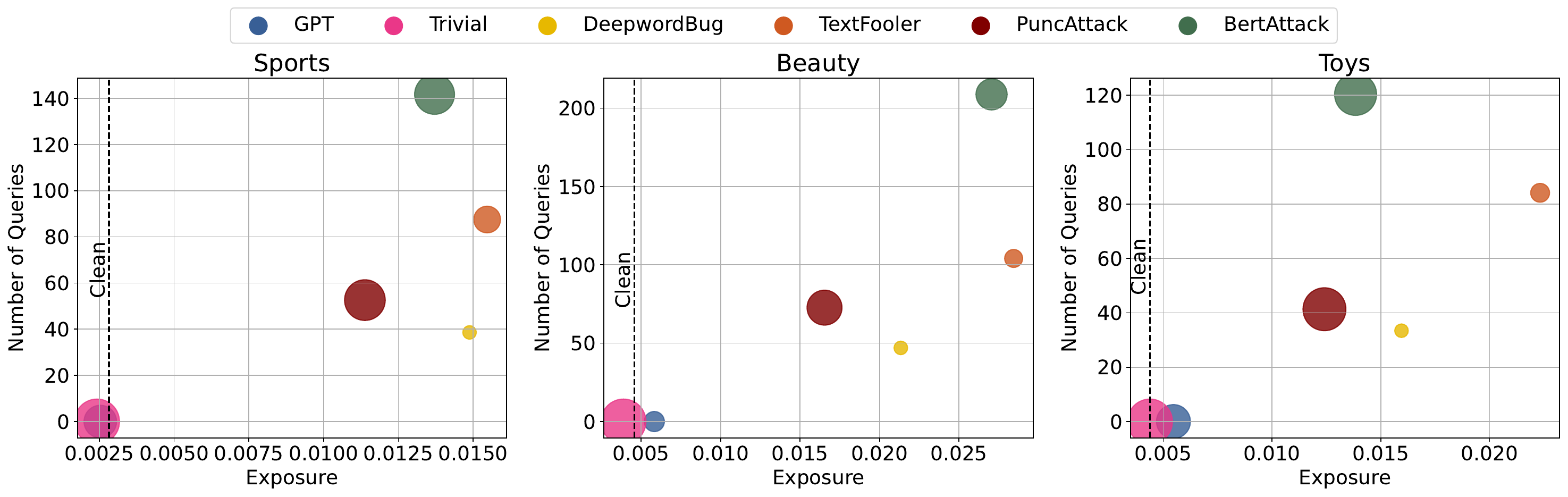}
    \caption{Performance comparison of different attacks on \textbf{RecFormer}. The size of the scatter points represents the cosine semantic similarity with the original title, with larger points indicating better semantic preservation (best viewed in color). }
    \label{fig:main}
\end{figure*}

\subsubsection{Datasets}
We conduct experiments on three categories of widely-used \cite{li2023Text, geng2022Recommendation, Bao2023TALLRecAE, zhang2023CoLLM} Amazon review dataset introduced by \citet{mcauley2015ImageBased}: `Beauty', `Toys and Games', 'Sports and Outdoors', which are named as \textbf{Beauty}, \textbf{Toys} and \textbf{Sports} in brief. We use the 5-core version of Amazon datasets where each user and item have 5 interactions at least. The statistics of these datasets are summarized in Appendix \ref{app:implement}.

\subsubsection{Implementation Details}
All victim models and compared shilling attacks are implemented in PyTorch. We random select 10\% items as target items. For more implementation details, please refer to Appendix \ref{app:implement}.

\subsubsection{Evaluation metrics}
We evaluate the attack from two aspects: \textit{effectiveness} and \textit{stealthiness}.

\textbf{Effectiveness}. This metric gauges the extent to which our methodology can promote the specified target items.
\begin{itemize}
    \item \textit{Exposure}.  For the victim model RecFormer, which allows for full ranking, we employed a direct metric, exposure rate. We define the exposure rate as $exp_i = \frac{N_{rec}^i}{N_{u}}$, where $N_{u}$ represents the total number of users, and $N_{rec}^i$ denotes the count of users for whom target item $i$ appears in their top-K ($K=50$ by default) recommendation list.
    \item \textit{Purchasing propensity}. For other three victim models which could not conduct full ranking, we define the purchasing propensity of item $i$ as $p_i = \mathbb{E}_{u \in \mathcal{U}} \, \hat{y_{ui}}$, where $\hat{y_{ui}}$ denotes the predicted probability that user $u$ tends to interact with item $i$.
    \item \textit{\# queries.} In a black-box scenario, the adversary need query the victim model in order to detect any alterations in the output logits. The lower the value, the more effective the attack will be.
\end{itemize}

\textbf{Stealthiness}. The stealthy attack aims to promote target items while maintaining imperceptibility, thereby avoiding detection by users and platforms. 
Therefore, the coherence and authenticity of our generated content are crucial to uphold.
\begin{itemize}
    \item \textit{Overall recommendation performance}. An ideal stealthy attack should keep the overall recommendation performance unchanged. The recommendation performance includes Recall@K, NDCG@K and AUC.
    \item \textit{Text quality.} The generated adversarial content should be of high quality that are acceptable to users. Firstly, it should be consistent with the corresponding item. We measure the \textbf{cosine semantic similarity} and \textbf{ROUGE scores} between the original content and adversarial content for this purpose. Secondly, the adversarial content itself should be readable. We assess the fluency of the adversarial title, measured by the \textbf{perplexity} of GPT-Neo \cite{gpt-neo}.
    \item \textit{\# perturbed words.} The number of words changed on an average to generate an adversarial content. The lower the value, the more imperceptible the attack will be.
\end{itemize}

\begin{table*}[]
\resizebox*{\linewidth}{!}{
\begin{tabular}{@{}cccccccccc@{}}
\toprule
\multirow{2}{*}[-0.25em]{Attacks} 
           & \multicolumn{3}{c}{Sports}     & \multicolumn{3}{c}{Beauty}     & \multicolumn{3}{c}{Toys}       \\ \cmidrule(l){2-4} \cmidrule(l){5-7} \cmidrule(l){8-10} 
           & NDCG@10 & Recall@10 & Exposure & NDCG@10 & Recall@10 & Exposure & NDCG@10 & Recall@10 & Exposure \\ \midrule 
Clean      & 0.01311 & 0.02811   & 0.00289  & 0.03066 & 0.06451   & 0.00449  & 0.03672 & 0.07712   & 0.00422  \\
Random     & \textbf{0.01234} & 0.02779   & 0.00295  & \textbf{0.03045} & 0.06254   & 0.00402  & 0.03447 & 0.07455   & 0.00422  \\
Bandwagon  & 0.01232 & 0.02779   & 0.00299  & 0.02914 & 0.05934   & 0.00421  & 0.03508 & 0.07583   & 0.00434  \\
Aush       & 0.01241 & 0.02740   & 0.00283  & 0.03010 & 0.06239   & 0.00430  & 0.03254 & 0.07336   & 0.00382  \\ 
LegUP      & 0.01219 & \textbf{0.02780}   & 0.00299  & 0.03029 & \textbf{0.06391}   & 0.00416  & 0.03411 & 0.07249   & 0.00423  \\
TextFooler & 0.01228 & \textbf{0.02780}   & \textbf{0.01074}  & 0.02926 & 0.06117   & \textbf{0.01886}  & \textbf{0.03596} & \textbf{0.07619}   & \textbf{0.01725}  \\ \bottomrule
\end{tabular}
}
\caption{Performance comparison with shilling attacks when RecFormer serves as victim model.}
\label{tab:shilling-recformer}
\end{table*}

\begin{figure*}[h]
    \centering
    \includegraphics[width=\textwidth]{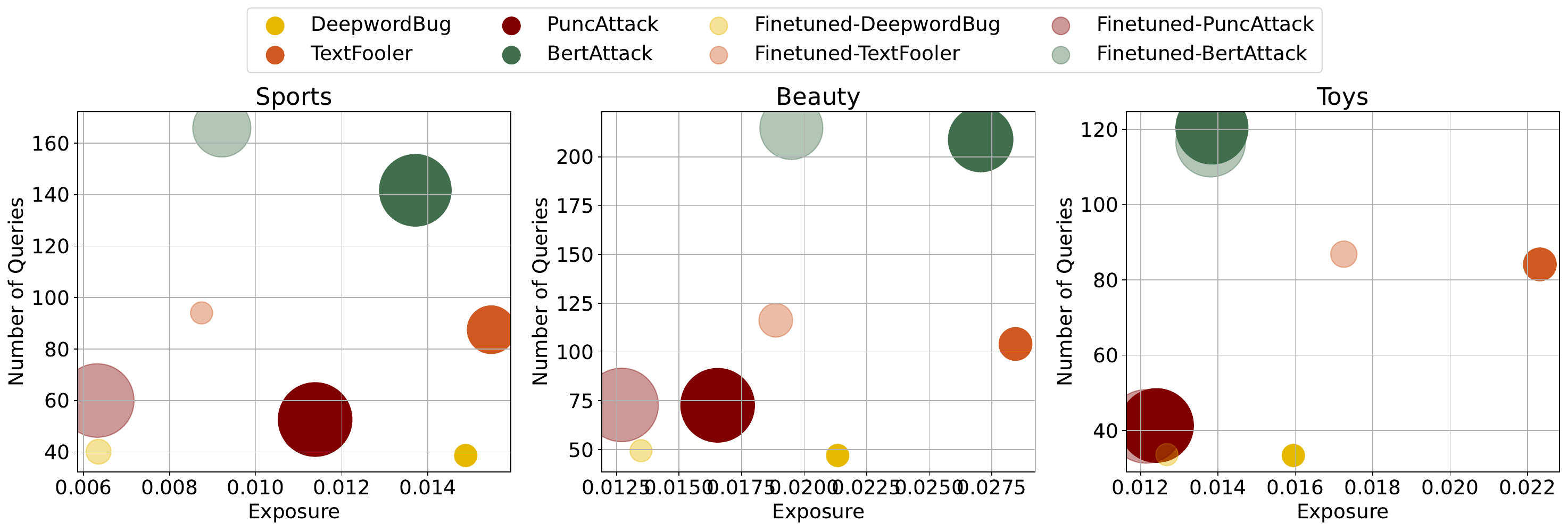}
    \caption{Performance comparison of different attacks on RecFormer. The size of the scatter points represents the cosine semantic similarity with the original title, with larger points indicating better semantic preservation.  (best viewed in color).}
    \label{fig:ft}
\end{figure*}

\subsection{Performance Comparison}
Table \ref{tab:recformer} shows the performance of all attacking methods on Recformer, P5, TALLRec and CoLLM, respectively.  The scatter plot and shilling attack comparison for RecFormer is in Figure \ref{fig:main} and Table \ref{tab:shilling-recformer}, while those for other victim models are in the appendix \ref{app:exp}. From them we can observe that:
\begin{itemize}
    \item \textbf{Our text attack paradigm can greatly promote the target items, demonstrating the vulnerability of LLM-based RS.} Even the simplest word insertion and rewriting using GPT can increase the exposure of the target item to a certain extent. Furthermore, black-box text attack methods could  
    lead to manifold increases in the exposure rate of the targeted items.
    \item \textbf{Our text attack strategy is also remarkably stealthy, making it difficult for users and platforms to detect.} Primarily, the overall performance of RS remains largely unchanged (even we choose 10\% items as target items), signifying that the attack does not disrupt the normal operation of RS. Additionally, the generated adversarial titles exhibit high semantic integrity that they are acceptable (or imperceptible) to human comprehension.
    \item \textbf{Traditional shilling attacks are not effective in LLM-based recommendation models.} Even with access to a portion of the training data, they fail to significantly enhance the exposure of the targeted items. This is attributed to the fact that LLM-based recommendation models prioritize content information primarily in textual form. Additionally, since fake user-generated training data is introduced during the model training phase, they significantly impact the overall performance of the victim model, which is easily detectable.
    \item  Word-level attacks are effective in boosting the exposure of target items, albeit demanding more queries and higher costs. On the other hand, character-level text attacks demonstrate superior results compared to shilling attacks, even with a reduced number of queries directed towards victim models.
\end{itemize}

\subsection{The influence of fine-tuning}
In this subsection, we examined the impact of model fine-tuning on attacks. A significant advantage of LLM-based recommendation models is their zero-shot transferability across datasets; however, we have also uncovered the vulnerability of such zero-shot models.

Figure \ref{fig:ft} shows a direct comparison between zero-shot RecFormer and fine-tuned Recformer. The detailed performance of fine-tuned Recformer is shown in Appendix \ref{app:exp}. From these, we observe that \textbf{fine-tuned models are more resilient to attacks compared to zero-shot models}. This is manifested in three aspects: attacking fine-tuned models requires more queries, yields lesser promoting of target items, and maintains poorer semantic integrity.

\subsection{The influence of item popularity}
In this section, we examined the impact of the initial popularity of target items on attacks. Popularity bias in recommendation systems favors popular items over personalized ones, limiting diversity, fairness and potentially dissatisfying users' preferences \cite{zhang2021causal}. We selected the top 150 and bottom 150 items in popularity from the dataset as target items and presented the attack results in Table \ref{tab:pop}. We can observe that \textbf{items with high popularity experience greater promotion, thereby exacerbating popularity bias}. High popularity target items can achieve greater exposure boosts with fewer queries and higher semantic consistency.

\begin{table}[h]
\resizebox*{\linewidth}{!}{
\begin{tabular}{@{}cccccccc@{}}
\toprule
\multirow{2}{*}{Dataset} & \multirow{2}{*}{Attack} & \multicolumn{2}{c}{Improved Exp. $\uparrow$} & \multicolumn{2}{c}{\# Queries $\downarrow$}  & \multicolumn{2}{c}{Cos. $\uparrow$}      \\ 
                         &                         & High                   & Low          & High           & Low            & High          & Low           \\ \midrule
\multirow{4}{*}{Toys}    & Deepwordbug             & \textbf{0.013}         & 0.007        & \textbf{28.9}  & 33.9           & 0.67          & \textbf{0.72} \\
                         & TextFooler              & \textbf{0.016}         & 0.013        & \textbf{71.1}  & 97.1           & \textbf{0.71} & 0.70          \\
                         & PuncAttack              & \textbf{0.012}         & 0.008        & \textbf{34.8}  & 47.5           & 0.84          & \textbf{0.85} \\
                         & BertAttack              & \textbf{0.011}         & 0.008        & \textbf{96.8}  & 107.8          & \textbf{0.84} & 0.83          \\ \midrule
\multirow{4}{*}{Beauty}  & Deepwordbug             & \textbf{0.014}         & 0.009        & \textbf{45.9}  & 58.3           & \textbf{0.79} & 0.71          \\
                         & TextFooler              & \textbf{0.024}         & 0.015        & 151.4          & \textbf{110.3} & \textbf{0.80} & 0.75          \\
                         & PuncAttack              & \textbf{0.013}         & 0.008        & 94.7           & \textbf{68.5}  & \textbf{0.87} & 0.85          \\
                         & BertAttack              & \textbf{0.022}         & 0.018        & 211.8          & \textbf{171.0} & \textbf{0.83} & 0.81          \\ \midrule
\multirow{4}{*}{Sports}  & Deepwordbug             & \textbf{0.005}         & 0.004        & \textbf{36.8}  & 39.7           & \textbf{0.79} & 0.77          \\
                         & TextFooler              & \textbf{0.008}         & 0.005        & \textbf{82.5}  & 101.6          & \textbf{0.82} & 0.72          \\
                         & PuncAttack              & \textbf{0.005}         & 0.003        & \textbf{54.2}  & 74.5           & 0.85          & \textbf{0.87} \\
                         & BertAttack              & \textbf{0.007}         & 0.005        & \textbf{147.1} & 161.3          & \textbf{0.83} & 0.82          \\ \bottomrule
\end{tabular}}
\caption{Performance comparison of target items with different popularity.}
\label{tab:pop}
\end{table}

\subsection{Transferability}
\subsubsection{Transferability across tasks.}
A significant feature of LLM-based recommendation models, like P5, is their capability to unify various recommendation tasks in a shared instruction-based framework. We evaluate the transferability of adversarial content across direct recommendation task and rating prediction task of P5. The results are presented in Table \ref{tab:transfer-tasks}. We can observe \textbf{there exists strong transferability between different tasks in such unified model}. Attacks targeting a single task can boost the exposure of target items across multiple tasks.
\begin{table}[h]
\centering
\resizebox*{0.9\linewidth}{!}{
\begin{tabular}{cccc} 
\toprule
            & Sports  & Beauty  & Toys     \\ \midrule
Clean       & 0.42900 & 0.28336 & 0.37671  \\
DeepwordBug & 0.43947 & 0.31062 & 0.39512  \\
TextFooler  & 0.45076 & 0.31238 & 0.41719  \\
PunAttack   & 0.43502 & 0.30748 & 0.38391  \\
BertAttack  & 0.43350 & 0.32314 & 0.39779  \\
\bottomrule
\end{tabular}
}
\caption{Results of user propensity scores in transferability experiments on the P5 model. Attack on direct recommendation task and apply the obtained adversarial text to sequence recommendation task.} 
\label{tab:transfer-tasks}
\end{table}

\subsubsection{Transferability across victim models.}
Firstly, we evaluate the transferability of the generated adversarial content across RecFormer, TALLRec and CoLLM. We select one model as the source model and apply the adversarial content generated from attacking it to two other models to verify if it can similarly boost target items. Experimental results of TextFooler are presented in Table \ref{tab:transfer-models} and similar trends are observed with other attack methods as well. We observe that \textbf{transferability only exists among recommendation models utilizing the same backbone LLMs}: there's strong mutual transferability between CoLLM and TALLRec because both models are based on LLaMA \cite{touvron2023llama} as the backbone; whereas, there is no transferability between Recformer (using Longformer \cite{Beltagy2020Longformer} as the backbone) and either of the former two.

\begin{table*}[h]
\centering
\small
\begin{tabular}{@{}cccccccc@{}}
\toprule
\multirow{2}{*}[-0.5em]{Source} & \multirow{2}{*}[-0.5em]{Target}    & \multicolumn{2}{c}{Sports}                           & \multicolumn{2}{c}{Beauty}                           & \multicolumn{2}{c}{Toys}    \\ \cmidrule(l){3-4}  \cmidrule(l){5-6} \cmidrule(l){7-8}
                        &                            & Ori Exp.                          & Att Exp.            & Ori Exp.                          & Att Exp.            & Ori Exp.                          & Att Exp.            \\  \midrule
TALLRec                 & \multirow{2}{*}{RecFormer} & \multirow{2}{*}{\textbf{0.00198}} & 0.00086          & \multirow{2}{*}{\textbf{0.00144}} & 0.00040          & \multirow{2}{*}{\textbf{0.00408}} & 0.00255          \\
CoLLM                   &                            &                                   & 0.00103          &                                   & 0.00064          &                                   & 0.00187          \\ \midrule
RecFormer               & \multirow{2}{*}{TALLRec}   & \multirow{2}{*}{0.10430}          & 0.08641          & \multirow{2}{*}{0.15917}          & 0.14881          & \multirow{2}{*}{0.67980}          & 0.65890          \\
CoLLM                   &                            &                                   & \textbf{0.12237} &                                   & \textbf{0.21140} &                                   & \textbf{0.72888} \\ \midrule
RecFormer               & \multirow{2}{*}{CoLLM}     & \multirow{2}{*}{0.58699}          & 0.58001          & \multirow{2}{*}{0.21619}          & 0.20556          & \multirow{2}{*}{0.35964}          & 0.33338          \\
TALLRec                 &                            &                                   & \textbf{0.62473} &                                   & \textbf{0.27289} &                                   & \textbf{0.39666} \\ \bottomrule
\end{tabular}
\caption{The results of transfer attack across different victim models when using TextFooler as the attack model. The best result is in \textbf{boldface}.} 
\label{tab:transfer-models}
\end{table*}

\begin{table*}[]
\resizebox*{\textwidth}{!}{
\begin{tabular}{@{}cccccccccccccc@{}}
\toprule
                           &                          & \multicolumn{3}{c}{DeepwordBug}                                                                     & \multicolumn{3}{c}{PunAttack}                                                                       & \multicolumn{3}{c}{TextFooler}                                                                      & \multicolumn{3}{c}{BertAttack}                                                                      \\ \cmidrule(l){3-5}  \cmidrule(l){6-8} \cmidrule(l){9-11} \cmidrule(l){12-14}  
\multirow{-2}{*}[0.25em]{Metrics}  & \multirow{-2}{*}[0.25em]{Attack} & Sports                          & Beauty                          & Toys                            & Sports                          & Beauty                          & Toys                            & Sports                          & Beauty                          & Toys                            & Sports                          & Beauty                          & Toys                            \\ \midrule
                           & Clean                    & 0.00282                         & 0.00458                         & 0.00439                         & 0.00282                         & 0.00458                         & 0.00439                         & 0.00282                         & 0.00458                         & 0.00439                         & 0.00282                         & 0.00458                         & 0.00439                         \\
                           & Attack                   & \cellcolor[HTML]{FFCCC9}0.01488 & \cellcolor[HTML]{FFCCC9}0.02134 & \cellcolor[HTML]{FFCCC9}0.01595 & \cellcolor[HTML]{FFCCC9}0.01138 & \cellcolor[HTML]{FFCCC9}0.01654 & \cellcolor[HTML]{FFCCC9}0.01241 & \cellcolor[HTML]{96FFFB}0.01547 & \cellcolor[HTML]{96FFFB}0.02844 & \cellcolor[HTML]{96FFFB}0.02232 & \cellcolor[HTML]{96FFFB}0.01371 & \cellcolor[HTML]{96FFFB}0.02705 & \cellcolor[HTML]{96FFFB}0.01384 \\ 
\multirow{-3}{*}{Exposure} & Defense                  & \cellcolor[HTML]{FFCCC9}0.00349 & \cellcolor[HTML]{FFCCC9}0.00587 & \cellcolor[HTML]{FFCCC9}0.00551 & \cellcolor[HTML]{FFCCC9}0.00399 & \cellcolor[HTML]{FFCCC9}0.00601 & \cellcolor[HTML]{FFCCC9}0.00623 & \cellcolor[HTML]{96FFFB}0.01510 & \cellcolor[HTML]{96FFFB}0.02012 & \cellcolor[HTML]{96FFFB}0.01867        & \cellcolor[HTML]{96FFFB}0.01065        & \cellcolor[HTML]{96FFFB}0.02043 & \cellcolor[HTML]{96FFFB}0.01161 \\ \midrule
                           & Clean                    & 0.00780                         & 0.01258                         & 0.02380                         & 0.00780                         & 0.01258                         & 0.02380                         & 0.00780                         & 0.01258                         & 0.02380                         & 0.00780                         & 0.01258                         & 0.02380                         \\
                           & Attack                   & 0.00757                         & 0.01257                         & 0.02365                         & 0.00762                         & 0.01257                         & 0.02328                         & 0.00780                         & 0.01224                         & 0.02366                         & 0.00781                         & 0.01213                         & 0.02340                         \\
\multirow{-3}{*}{NDCG@10}  & Defense                  & 0.00769                         & 0.01251                         & 0.02373                         & 0.00771                         & 0.01251                         & 0.02377                         & 0.00778                         & 0.01217                         & 0.02378                         & 0.00774                         & 0.01195                         & 0.02351                         \\ \bottomrule
\end{tabular}
}
\caption{Defense performance on RecFormer. The red highlighted area indicates effective defense against character-level attacks, while the blue highlighted area indicates limited defense against word-level attacks.}
\label{tab:defense-recformer}
\end{table*}

\subsection{Re-writing Defense}
\label{sec:defense}
In this subsection, we explore potential strategies for addressing the identified vulnerability of LLM-based recommendation models. The most direct strategy is to detect and revise potential adversarial elements in the content, such as spelling errors and potential word substitutions. We utilize GPT-3.5-turbo to accomplish the rewriting of adversarial content to achieve defense. 
\begin{center}
    \fcolorbox{black}{gray!20}{ \parbox{0.9\linewidth}{
    \textbf{Re-writing Prompt}: Correct possible grammar, spelling and word substitution errors in the product title (dirctly output the revised title only): <AdversarialTitle>}}
\end{center}
The exposure rates and recommendation performance (NDCG@10) before and after defense on RecFormer are shown in Table \ref{tab:defense-recformer}. The results of other victim models are shown in Appendix \ref{app:exp}. We can observe\textbf{ the defense works well against character-level attacks} like DeepwordBug and PuncAttack, \textbf{but struggles with more complex word substitution attacks} such as TextFooler and BertAttack, since character-level spelling errors and punctuation insertions are relatively easy to detect. Moreover, it doesn't impact overall recommendation performance.

\section{Related Work}

\subsection{LLM-based Recommendation}
The techniques used by LLMs in the recommendation domain \cite{li2024survey,bao2023large,lin2023can} involve translating recommendation tasks into natural language tasks and adapting LLMs to generate recommendation results directly. These generative LLM-based approaches can be further divided into two paradigms based on whether parameters are tuned: non-tuning and tuning paradigms. The non-tuning paradigm assumes LLMs already have the recommendation abilities and attempt to trigger the strong zero/few-shot abilities by introducing specific prompts~\cite{Liu2023IsCA,Dai2023UncoveringCC,Mysore2023LargeLM,Wang2023RethinkingTE,Hou2023LargeLM,wang2024learnable}.
The tuning paradigm uses fine-tuning, prompt learning, or instruction tuning~\cite{Kang2023DoLU,Bao2023TALLRecAE,Wang2022TowardsUC,geng2022Recommendation,cui2022M6Rec} to enhance LLM's recommendation abilities by using LLMs as encoders to extract user and item representations and then fine-tuning their parameters on specific loss functions.

\subsection{Shilling Attack}
Shilling attacks aim to interfere with the recommendation strategy of a victim recommender system by injecting fake users into the training matrix~\cite{Deldjoo2019AssessingTI,Toyer2023TensorTI,wang2024uplift}. This can be implemented through (1) heuristic attacks~\cite{Burke2005LimitedKS,Linden2003AmazoncomRI,kaur2016Shilling}, where fake profiles are created based on subjective inference and existing knowledge; (2) gradient attacks~\cite{Fang2020InfluenceFB,Li2016DataPA, Zhang2020PracticalDP,Fang2018PoisoningAT,Huang2021DataPA}, which optimize the objective function through a continuous space; and neural attacks ~\cite{Wang2023PredictionOT,lin2020Attacking,lin2024ShillingBlackbox,Song2020PoisonRecAA,Zhang2022TargetedDP}, which use deep learning to generate realistic profiles.

\section{Conclusion}
In conclusion, our investigation exposes a critical security issue within LLM-based recommendation systems, brought on by their reliance on textual content. By showcasing the ability of attackers to boost item exposure through subtle text modifications, we stress the urgent need for heightened security measures. Our findings not only highlight the vulnerability of these systems but also serve as a call to action for the development of more robust, attack-resistant models.
\section*{Limitations}
The main constraints can be summarized in the following two aspects: Firstly, although the black-box text attack model does not require access to the victim model's parameters and gradients, it necessitates multiple queries to the model. And it is challenging to query the model in real-world large-scale recommendation systems \cite{ying2018graph}. Secondly, this study solely focuses on the content features of text modality. In reality, recommendation systems encompass other modalities such as images and videos \cite{wei2019mmgcn,zhang2021mining,zhang2022latent,zhang2023mining}. The issue of attacking models based on these modalities also represents a worthy direction for research. 

\section*{Ethics Statement}
The experimental datasets are publicly available from some previous works, downloaded via official APIs. The information regarding users in the Amazon dataset has been anonymized, ensuring there are no privacy concerns related to the users. We do not disclose any non-open-source data, and we ensure that our actions comply with ethical standards. We use publicly available pre-trained models, i.e., RecFormer, P5, LLaMA, GPT-Neo. All the checkpoints and datasets are carefully processed by their authors to ensure that there are no ethical problems. 

However, it is worth noting that our research has uncovered vulnerabilities in LLM-based RS. Despite proposing potential defense methods in Section \ref{sec:defense}, there still exists a risk of misuse of our attack paradigm. Future research based on this attack should proceed with caution and consider the potential consequences of any proposed methods.

\section*{Acknowledgements}
This work is supported by National Natural Science Foundation of China (62141608, 62206291, 62236010).

\bibliography{acl}

\appendix
\section{Experimental Settings}
\label{sec:appendix}
\subsection{Victim Models}
\label{app:victim}
We choose four mainstream LLM-based recommendation models as our victim models.
\begin{itemize}
    \item \textbf{Recformer} \cite{li2023Text}. Recformer proposes to formulate an item as a "sentence" (word sequence) and can effectively recommend the next item based on language representations.
    \item \textbf{P5} \cite{geng2022Recommendation}. P5 presents a flexible and unified text-to-text paradigm called "Pretrain, Personalized Prompt, and Predict Paradigm" (P5) for recommendation, which unifies various recommendation tasks in a shared framework.
    \item \textbf{TALLRec} \cite{Bao2023TALLRecAE}. TALLRec proposes to align LLMs with recommendation by tunning LLMs with recommendation data.
    \item \textbf{CoLLM} \cite{zhang2023CoLLM}. CoLLM seamlessly incorporates collaborative information into LLMs for recommendation by mapping ID embedding to the input token embedding space of LLM.
\end{itemize}

\begin{table}[h]
    \caption{Statistics of the datasets}
    \resizebox*{\linewidth}{!}{
    \begin{tabular}{ccccc}
    \toprule
    Dataset  & \#Users & \#Items & \#Interactions & Density \\
    \midrule
    Sports   & 35,598   & 18,357   & 256,308         & 0.00039 \\
    Beauty   & 22,363   & 12,101   & 172,188         & 0.00064 \\
    Toys     & 19,412   & 11,924   & 145.004         & 0.00063 \\
    \bottomrule
    \end{tabular}}
    \label{tab:dataset}
\end{table}

\subsection{Compared Shilling Attacks}
\label{app:baselines}
Shilling attacks aim to generate fake users that assign high ratings for a target item, while also rating other items to act like normal users for evading. 
\begin{itemize}
    \item Heuristic attacks. Heuristic attacks involve selecting items to create fake profiles based on heuristic rules. \textbf{Random attack} \cite{kaur2016Shilling} selects filler items randomly while \textbf{Bandwagon attack} \cite{burke2005Segmentbased} selects the popular items as users fake preferences, which is white-box as it requires knowledge of the popularity of items, i.e., the training data.
    \item Neural attacks. Neural attacks utilize neural networks to generate fake users that maximize the objective function. \textbf{Aush} \cite{lin2020Attacking}. Aush utilizes Generative Adversarial Network (GAN) to generate fake users based on known knowledge. \textbf{Leg-UP} \cite{lin2024ShillingBlackbox}. Leg-UP learns user behavior patterns from real users in the sampled “templates” and constructs fake user profiles. Both of them are gray-box attack models, requiring a portion of training data.
\end{itemize}

\subsection{Implementation Details}
\label{app:implement}
The statistics of these datasets are summarized in Table \ref{tab:dataset}. All victim models and compared shilling attacks are implemented in PyTorch. We random select 10\% items as target items for each dataset. For RecFormer, we use both pre-trained checkpoint\footnote{https://github.com/AaronHeee/RecFormer} and also fine-tune it with the three datasets. For P5, we directly use the fine-tuned checkpoints\footnote{https://github.com/jeykigung/P5}. For TALLRec\footnote{https://github.com/SAI990323/TALLRec} and CoLLM\footnote{https://github.com/zyang1580/CoLLM}, we fine-tune them from scratch. We implement shilling attack methods using RecAD\footnote{https://github.com/gusye1234/recad}. Both Aush and Leg-UP are gray-box and we set them to access 20\% of the training data.

\section{Text Attack}
\label{app:attack}
\subsection{Text Attack Components}
A textual attack consists of four main components: \textit{Goal Function}, \textit{Constraint}, \textit{Transformation}, and \textit{Search Method}. Here is a breakdown example of each component.

\subsubsection{Goal Function}
The Goal Function defines the objective of the attack. It also scores how ”good” the given manipulated text is for achieving the desired goal. The core part could be simplified as:

\begin{lstlisting}[language=Python]
def goal_function(target_item_id, original_text, perturbed_text, threshold=0.5):
    """
    Return the attacked score and determines if the attack is successful.
    
    :param target_item_id: The target item's id.
    :param original_text: The original text of target item.
    :param perturbed_text: The perturbed text of target item.
    :param threshold: The threshold of success attack.
    :return: attacked_score, is_successful
    """
    init_score = call_model(original_text, target_item_id)
    attacked_score = call_model(perturbed_text, target_item_id)
    is_successful = attacked_score - init_score > threshold
    return attacked_score, is_successful
\end{lstlisting}

\subsubsection{Constraint}
Constraints are conditions that must be met for the perturbed text to be considered valid. These often ensure the perturbed text remains natural and similar to the original text in some aspects (e.g., semantic similarity). Examples are as follows:

\begin{lstlisting}[language=Python]
def maintain_semantic(original_text, perturbed_text, threshold=0.8):
    """
    Checks if the perturbed text maintains semantic similarity.
    
    :param original_text: Original text.
    :param perturbed_text: Perturbed version of the text.
    :param threshold: Threshold for semantic similarity.
    :return: True if similarity is above the threshold, False otherwise.
    """
    similarity = compute_semantic_similarity(input_text, perturbed_text)
    return similarity > threshold
\end{lstlisting}

\subsubsection{Transformation}
The Transformation component refers to the methods applied to modify the original text to achieve the adversarial goal. This could involve synonym replacement, insertion, or deletion of words.

    

\begin{lstlisting}[language=Python]
def synonym_replacement(original_text):
    """
    Manipulates the original text by replacing synonyms.

    :param original_text: The original text to be manipulated.
    :return: A list of manipulated texts.
    """
    words = original_text.words
    transformed_texts = []
    for i in range(len(words)):
        replacement_word = get_synonyms(words[i])
        modified_text = original_text.replace_word_at_index(i, replacement_word)
        transformed_texts.append(modified_text)
    return transformed_texts
\end{lstlisting}
    
\subsubsection{Search Method}
The Search Method dictates the strategy used to explore the space of possible perturbations. For example, a greedy search method might iteratively apply transformations that maximally increase the attack's success likelihood.

\begin{lstlisting}[language=Python] 
def greedy_search(target_item_id, original_text):
    """
    Applies greedy search to find successful perturbation.
    
    :param target_item_id: The target item's id.
    :param original_text: Original text to be perturbed.
    :return: Best perturbed text.
    """
    best_score = 0
    best_perturbed_text = original_text
    perturbed_texts = get_transformations(original_text)
    for perturbed_text in perturbed_texts:
        attacked_score, is_successful = goal_function(target_item_id, original_text, perturbed_text)
        if satisfy_constraints(original_text, perturbed_text):
            if attacked_score > best_score:
                best_score = attacked_score
                best_perturbed_text = perturbed_text
                if is_successful:
                    return best_perturbed_text
    return None
\end{lstlisting}

\subsection{Implementations}
The majority of our text attacks have been developed by revising strategies from TextAttack\footnote{https://github.com/QData/TextAttack} \cite{morris2020TextAttack} and PromptBench\footnote{https://github.com/microsoft/promptbench} \cite{zhu2023PromptBench}.

All four attack methods, DeepwordBug, PuncAttack, TextFooler and BertAttack share the same \textit{Goal Function}, where we set the success threshold to $0.05$ increasing exposure rate for RecFormer. Since the other three victim models cannot perform full ranking and calculate exposure rates, we set an increase in interaction probability as the objective. Specifically, we set success threshold to $0.3, 0.15, 0.15$ increasing interaction probability for P5, TALLRec and CoLLM, respectively. During the attack process, we randomly select 10\% of users to calculate the average exposure rate or interaction probability instead of using all users. This approach is lower in cost and more aligned with the constraints of practical attacks.

The recipes of \textit{Constraint}, \textit{Transformation}, and \textit{Search Method} of implemented text attacks are as follows:

\begin{lstlisting}[language=Python]
"""
Recipes for DeepwordBug
"""

transformation = CompositeTransformation(
    [
        WordSwapNeighboringCharacterSwap(),
        WordSwapRandomCharacterSubstitution(),
        WordSwapRandomCharacterDeletion(),
        WordSwapRandomCharacterInsertion(),
    ]
)

constraints = [
    RepeatModification(), 
    StopwordModification(),
    LevenshteinEditDistance(30)
    ]
search_method = GreedyWordSwapWIR() 
\end{lstlisting}

\begin{lstlisting}[language=Python]
"""
Recipes for PuncAttack
"""
punctuations = '\'-'
transformation = WordSwapTokenSpecificPunctuationInsertion(letters_to_insert=punctuations)
constraints = [
    RepeatModification(), 
    StopwordModification(),
    WordEmbeddingDistance(min_cos_sim=0.6),
    PartOfSpeech(allow_verb_noun_swap=True),
    UniversalSentenceEncoder(threshold=0.8)
    ]
search_method =  GreedyWordSwapWIR()
\end{lstlisting}

\begin{lstlisting}[language=Python]
"""
Recipes for TextFooler
"""
transformation = WordSwapEmbedding(max_candidates=50)
constraints = [
    RepeatModification(), 
    StopwordModification(),
    WordEmbeddingDistance(min_cos_sim=0.6),
    PartOfSpeech(allow_verb_noun_swap=True),
    UniversalSentenceEncoder(threshold=0.84,metric="angular")
    ]
search_method = GreedyWordSwapWIR()
\end{lstlisting}

\begin{lstlisting}[language=Python]
"""
Recipes for BertAttack
"""
transformation = WordSwapMaskedLM(max_candidates=48)
constraints = [
    RepeatModification(), 
    StopwordModification(),
    MaxWordsPerturbed(max_percent=1),
    UniversalSentenceEncoder(threshold=0.8)
    ]
search_method = GreedyWordSwapWIR()
\end{lstlisting}

\section{Detailed Experimental Results}
\label{app:exp}
\begin{figure*}[]
    \centering
    \begin{subfigure}
        \centering
        \includegraphics[width=\textwidth]{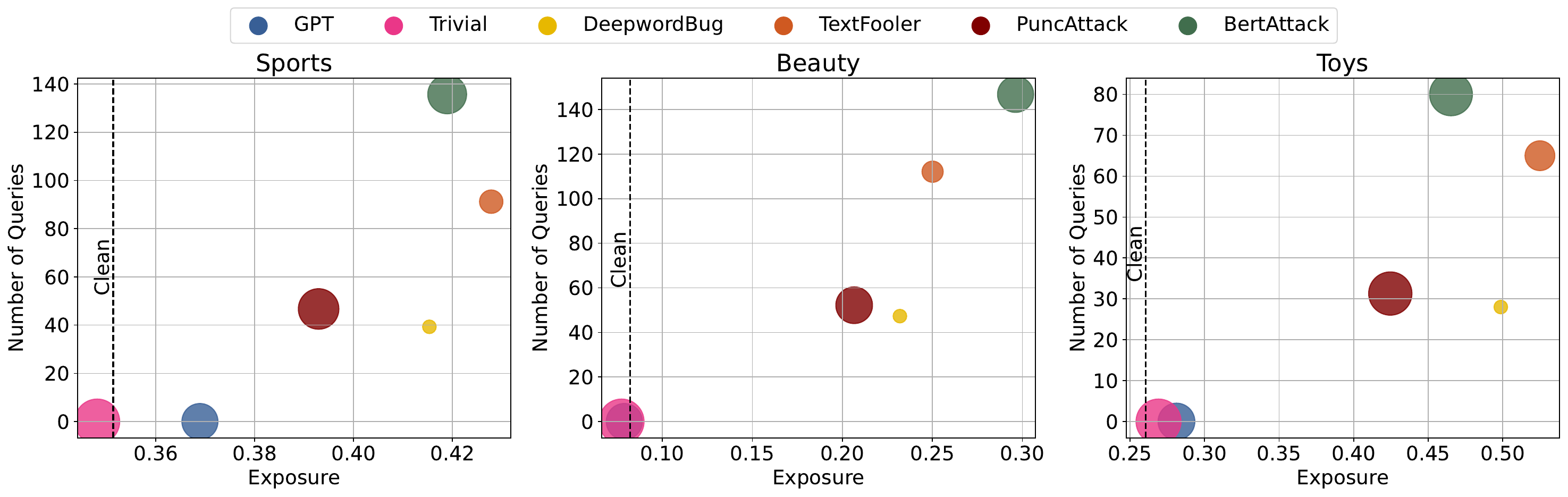}
        \caption{Performance comparison of different attacks on \textbf{P5}.}
    \end{subfigure}
    \begin{subfigure}
        \centering
        \includegraphics[width=\textwidth]{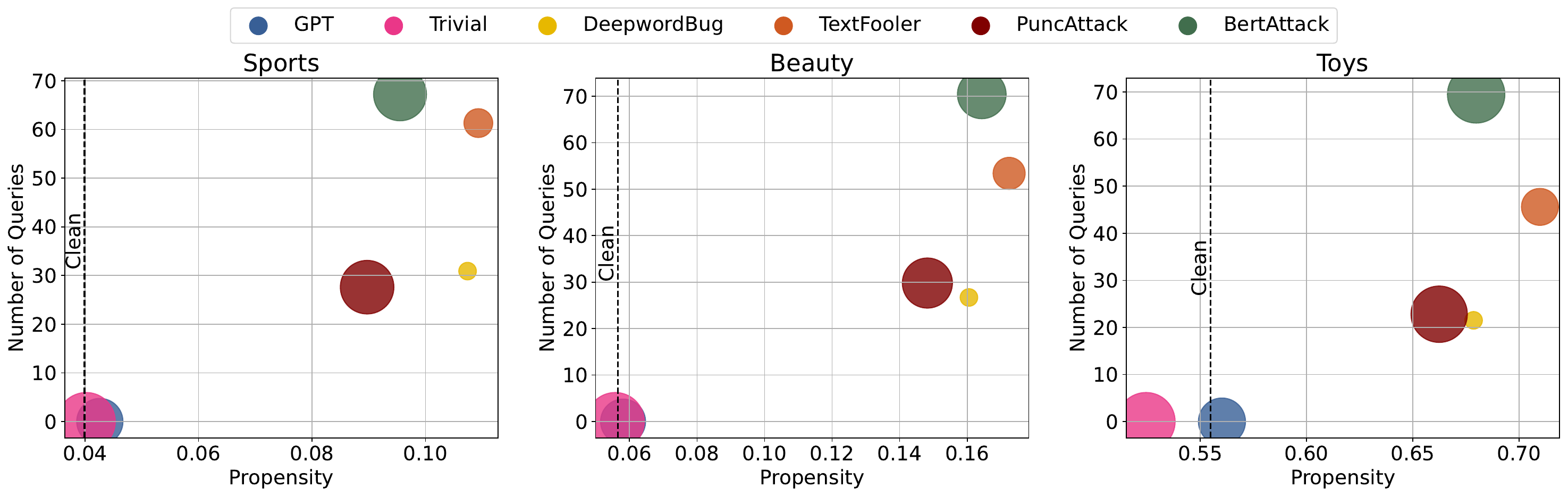}
        \caption{Performance comparison of different attacks on \textbf{TALLRec}.}
    \end{subfigure}
    \begin{subfigure}
        \centering
        \includegraphics[width=\textwidth]{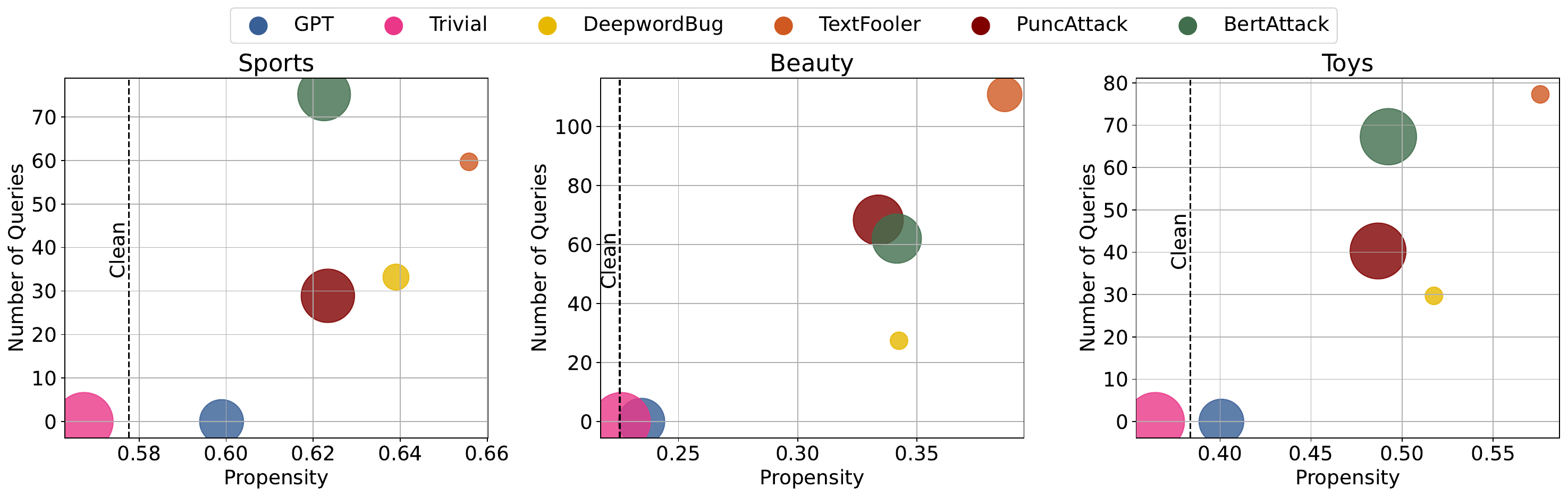}
        \caption{Performance comparison of different attacks on \textbf{CoLLM}.}
    \end{subfigure}
    \caption{Performance comparison of different attacks on various models. The size of the scatter points represents the cosine semantic similarity with the original title, with larger points indicating better semantic preservation (best viewed in color).}
    \label{fig:scatters}
\end{figure*}

In this section, we present detailed experimental results that could not be shown in the main text due to space limitation. 
\begin{itemize}
    \item The overall attack performances of P5, TALLRec and CoLLM are shown in Table \ref{tab:p5}, Table \ref{tab:tallrec} and Table \ref{tab:collm}, respectively. The scatter plots of them are shown in Figure \ref{fig:scatters}.
    \item The performances of attacking fine-tuned Recformer is shown in Table \ref{tab:recformer-ft}.
    \item The performance comparison with traditional shilling attacks of TALLRec and CoLLM are shown in \ref{tab:shilling-tallrec} and  \ref{tab:shilling-collm}.
    \item We supplemented the experimental results on the MIND-small news recommendation dataset \cite{wu-etal-2020-mind} with RecFormer as victim model. The performance is shown in Table \ref{tab:mind}. It can be observed that in the news scenario, LLM-based recommendation systems are also vulnerable to text attacks. This could potentially boost the dissemination of low-quality, fake news, or news with evident bias.
    \item We conducted experiments on the impact of different numbers of fake users on the effectiveness of shilling attacks in LLM-based recommendation systems. The results are shown in Table \ref{tab:fakeusers}. We can observe that as the number of fake users increases, the exposure of the target items also improves. However, compared to the effects of text attacks, this improvement is limited. Additionally, shilling attack shows less stealthiness. As the number of fake users increases, there is a noticeable decline in the overall performance of the recommendation model, which could make it easier for platforms to detect such attacks.

\end{itemize}

\begin{table}[h]
    \centering
    \resizebox*{\linewidth}{!}{
    \begin{tabular}{cccc}
        \toprule
        Method & Exposure & Rel. Impro. & \# queries \\
        \midrule
        Clean & 0.00723 & - & - \\
        DeepwordBug & 0.02655 & 267.2\% & 39.6 \\
        TextFooler & 0.04561 & 530.8\% & 101.7 \\
        PuncAttack & 0.02712 & 275.1\% & 60.1 \\
        BertAttack & 0.04178 & 477.9\% & 104.3 \\
        \bottomrule
    \end{tabular}}
    \caption{Performance comparison of attacking \textbf{finetuned Recformer} on MIND-small dataset.}
    \label{tab:mind}
\end{table}

\begin{table}[h]
    \centering
    \resizebox*{\linewidth}{!}{
    \begin{tabular}{ccc}
        \toprule
        Number of fake users & Exposure & NDCG@10 \\
        \midrule
        0 (Clean) & 0.00422 & 0.03672 \\
        100 & 0.00434 & 0.03508 \\
        200 & 0.00448 & 0.03514 \\
        500 & 0.00463 & 0.03466 \\
        1000 & 0.00481 & 0.03403 \\
        TextFooler & 0.01886 & 0.03596 \\
        \bottomrule
    \end{tabular}}
    \caption{Comparison of different numbers of fake users of shilling attack method Bandwagon on Amazon-Toys dataset (\textbf{RecFormer} as victim model).}
    \label{tab:fakeusers}
\end{table}

\begin{table*}[]
\resizebox*{\textwidth}{!}{
\begin{tabular}{@{}cccccccccc@{}}
\toprule
\multicolumn{1}{l}{\multirow{2}{*}{Dataset}} & \multirow{2}{*}{Method}          & \multicolumn{3}{c}{Effectiveness}                                              & \multicolumn{5}{c}{Stealthiness}                                                                                    \\ \cmidrule(l){3-5} \cmidrule(l){6-10} 
\multicolumn{1}{l}{}                         &                                  & Propensity $\uparrow$ & Rel. Impro. $\uparrow$ & \# queries $\downarrow $           & NDCG@10 $\uparrow$ & Cos. $\uparrow$ & Rouge-l $\uparrow$ & Perplexity $\downarrow $ & \# pert. words $\downarrow $ \\ \midrule
\multicolumn{1}{c|}{\multirow{7}{*}{Sports}} & \multicolumn{1}{c|}{Clean}       & 0.35137             & 0.0\%               & \multicolumn{1}{c|}{-}            & 0.28782            & 1.000           & 1.000              & 2158.7                   & -                           \\
\multicolumn{1}{c|}{}                        & \multicolumn{1}{c|}{ChatGPT}     & 0.36891             & 5.0\%               & \multicolumn{1}{c|}{-}            & 0.28777            & 0.794           & 0.499              & \textbf{1770.9}          & -                           \\
\multicolumn{1}{c|}{}                        & \multicolumn{1}{c|}{Trivial}      & 0.34813             & -0.9\%              & \multicolumn{1}{c|}{-}            & 0.28764            & \textbf{0.896}  & \textbf{0.869}     & 4376.6                   & -                           \\
\multicolumn{1}{c|}{}                        & \multicolumn{1}{c|}{Deepwordbug} & 0.41533             & 18.2\%              & \multicolumn{1}{c|}{\textbf{39.3}} & 0.28765            & 0.637           & 0.381              & 7417.3                   & 4.9                          \\
\multicolumn{1}{c|}{}                        & \multicolumn{1}{c|}{TextFooler}  & \textbf{0.42784}    & \textbf{21.8\%}     & \multicolumn{1}{c|}{91.2}          & 0.28768            & 0.688           & 0.499              & 2494.9                   & 4.1                          \\
\multicolumn{1}{c|}{}                        & \multicolumn{1}{c|}{PuncAttack}  & 0.39292             & 11.8\%              & \multicolumn{1}{c|}{46.7}          & \textbf{0.28789}   & 0.842           & 0.602              & 2572.7                   & \textbf{3.4}                 \\
\multicolumn{1}{c|}{}                        & \multicolumn{1}{c|}{BertAttack}  & 0.41893             & 19.2\%              & \multicolumn{1}{c|}{135.7}         & 0.28781            & 0.823           & 0.598              & 9421.4                   & 3.7                          \\ \midrule
\multicolumn{1}{c|}{\multirow{7}{*}{Beauty}} & \multicolumn{1}{c|}{Clean}       & 0.08218             & 0.0\%               & \multicolumn{1}{c|}{-}            & 0.28765            & 1.000           & 1.000              & 611.6                    & -                           \\
\multicolumn{1}{c|}{}                        & \multicolumn{1}{c|}{ChatGPT}     & 0.07889             & -4.0\%              & \multicolumn{1}{c|}{-}            & 0.28733            & 0.822           & 0.516              & \textbf{501.8}           & -                           \\
\multicolumn{1}{c|}{}                        & \multicolumn{1}{c|}{Trivial}      & 0.07734             & -5.9\%              & \multicolumn{1}{c|}{-}            & \textbf{0.28761}   & \textbf{0.939}  & \textbf{0.901}     & 1189.5                   & -                           \\
\multicolumn{1}{c|}{}                        & \multicolumn{1}{c|}{Deepwordbug} & 0.23193             & 182.2\%             & \multicolumn{1}{c|}{\textbf{47.3}} & 0.28708            & 0.640           & 0.370              & 4590.1                   & 4.9                          \\
\multicolumn{1}{c|}{}                        & \multicolumn{1}{c|}{TextFooler}  & 0.25010             & 204.3\%             & \multicolumn{1}{c|}{112.1}         & 0.28691            & 0.683           & 0.460              & 1200.3                   & 4.2                          \\
\multicolumn{1}{c|}{}                        & \multicolumn{1}{c|}{PuncAttack}  & 0.20653             & 151.3\%             & \multicolumn{1}{c|}{52.2}          & 0.28693            & 0.828           & 0.594              & 1132.0                   & \textbf{3.4}                 \\
\multicolumn{1}{c|}{}                        & \multicolumn{1}{c|}{BertAttack}  & \textbf{0.29618}    & \textbf{260.4\%}    & \multicolumn{1}{c|}{146.8}         & 0.28676            & 0.821           & 0.585              & 2634.5                   & 3.5                          \\ \midrule
\multicolumn{1}{c|}{\multirow{7}{*}{Toys}}   & \multicolumn{1}{c|}{Clean}       & 0.26065             & 0.0\%               & \multicolumn{1}{c|}{-}            & 0.28587            & 1.000           & 1.000              & 4060.4                   & -                           \\
\multicolumn{1}{c|}{}                        & \multicolumn{1}{c|}{ChatGPT}     & 0.28115             & 7.9\%               & \multicolumn{1}{c|}{-}            & 0.28610            & 0.793           & 0.454              & \textbf{1967.1}          & -                           \\
\multicolumn{1}{c|}{}                        & \multicolumn{1}{c|}{Trivial}      & 0.26913             & 3.3\%               & \multicolumn{1}{c|}{-}            & 0.28614            & \textbf{0.880}  & \textbf{0.852}     & 7874.4                   & -                           \\
\multicolumn{1}{c|}{}                        & \multicolumn{1}{c|}{Deepwordbug} & 0.49867             & 91.3\%              & \multicolumn{1}{c|}{\textbf{28.0}} & 0.28619            & 0.642           & 0.490              & 8896.5                   & 3.5                          \\
\multicolumn{1}{c|}{}                        & \multicolumn{1}{c|}{TextFooler}  & \textbf{0.52492}    & \textbf{101.4\%}    & \multicolumn{1}{c|}{65.0}          & 0.28613            & 0.733           & 0.571              & 4413.1                   & 3.0                          \\
\multicolumn{1}{c|}{}                        & \multicolumn{1}{c|}{PuncAttack}  & 0.42457             & 62.9\%              & \multicolumn{1}{c|}{31.3}          & \textbf{0.28637}   & 0.860           & 0.666              & 4134.3                   & \textbf{2.4}                 \\
\multicolumn{1}{c|}{}                        & \multicolumn{1}{c|}{BertAttack}  & 0.46528             & 78.5\%              & \multicolumn{1}{c|}{80.0}          & 0.28597            & 0.855           & 0.690              & 9618.9                   & \textbf{2.4}                 \\ \bottomrule
\end{tabular}
}
\caption{Performance comparison of attacking \textbf{P5} where Rel. Impro. denotes relative improvement against clean setting. The best result is in \textbf{boldface}.}
\label{tab:p5}
\end{table*}

\begin{table*}[]
\resizebox*{\textwidth}{!}{
\begin{tabular}{@{}cccccccccc@{}}
\toprule
\multicolumn{1}{l}{\multirow{2}{*}{Dataset}} & \multirow{2}{*}{Method}          & \multicolumn{3}{c}{Effectiveness}                                              & \multicolumn{5}{c}{Stealthiness}                                                                                    \\ \cmidrule(l){3-5} \cmidrule(l){6-10} 
\multicolumn{1}{l}{}                         &                                  & Propensity $\uparrow$ & Rel. Impro. $\uparrow$ & \# queries $\downarrow $           & NDCG@10 $\uparrow$ & Cos. $\uparrow$ & Rouge-l $\uparrow$ & Perplexity $\downarrow $ & \# pert. words $\downarrow $ \\ \midrule
\multicolumn{1}{c|}{\multirow{7}{*}{Sports}} & \multicolumn{1}{c|}{Clean}       & 0.0399              & -                   & \multicolumn{1}{c|}{-}             & 0.58489            & 1.000           & 1.000              & 2158.7                   & -                            \\
\multicolumn{1}{c|}{}                        & \multicolumn{1}{c|}{ChatGPT}     & 0.0426              & 6.7\%               & \multicolumn{1}{c|}{-}             & 0.58449            & 0.794           & 0.499              & \textbf{1770.9}          & -                            \\
\multicolumn{1}{c|}{}                        & \multicolumn{1}{c|}{Trivial}      & 0.0402              & 0.6\%               & \multicolumn{1}{c|}{-}             & \textbf{0.58479}   & \textbf{0.896}  & \textbf{0.869}     & 4376.6                   & -                            \\
\multicolumn{1}{c|}{}                        & \multicolumn{1}{c|}{Deepwordbug} & 0.1074              & 168.8\%             & \multicolumn{1}{c|}{30.9}          & 0.58459            & 0.643           & 0.459              & 8274.9                   & 3.3                          \\
\multicolumn{1}{c|}{}                        & \multicolumn{1}{c|}{TextFooler}  & \textbf{0.1093}     & \textbf{173.5\%}    & \multicolumn{1}{c|}{61.3}          & 0.58395            & 0.686           & 0.512              & 2075.5                   & 2.7                          \\
\multicolumn{1}{c|}{}                        & \multicolumn{1}{c|}{PuncAttack}  & 0.0897              & 124.6\%             & \multicolumn{1}{c|}{\textbf{27.6}} & \textbf{0.58489}   & 0.854           & 0.621              & 3153.9                   & 2.0                          \\
\multicolumn{1}{c|}{}                        & \multicolumn{1}{c|}{BertAttack}  & 0.0955              & 139.1\%             & \multicolumn{1}{c|}{67.2}          & 0.58484            & 0.848           & 0.682              & 5397.2                   & \textbf{1.7}                 \\ \midrule
\multicolumn{1}{c|}{\multirow{7}{*}{Beauty}} & \multicolumn{1}{c|}{Clean}       & 0.0566              & -                   & \multicolumn{1}{c|}{-}             & 0.56758            & 1.000           & 1.000              & 611.6                    & -                            \\
\multicolumn{1}{c|}{}                        & \multicolumn{1}{c|}{ChatGPT}     & 0.0581              & 2.6\%               & \multicolumn{1}{c|}{-}             & 0.56546            & 0.822           & 0.516              & \textbf{501.8}           & -                            \\
\multicolumn{1}{c|}{}                        & \multicolumn{1}{c|}{Trivial}      & 0.0558              & -1.3\%              & \multicolumn{1}{c|}{-}             & \textbf{0.56755}   & \textbf{0.939}  & \textbf{0.901}     & 1189.5                   & -                            \\
\multicolumn{1}{c|}{}                        & \multicolumn{1}{c|}{Deepwordbug} & 0.1605              & 183.6\%             & \multicolumn{1}{c|}{\textbf{26.7}} & 0.56737            & 0.674           & 0.531              & 2330.8                   & 2.9                          \\
\multicolumn{1}{c|}{}                        & \multicolumn{1}{c|}{TextFooler}  & \textbf{0.1724}     & \textbf{204.6\%}    & \multicolumn{1}{c|}{53.4}          & 0.56581            & 0.736           & 0.589              & 1491.0                   & 2.5                          \\
\multicolumn{1}{c|}{}                        & \multicolumn{1}{c|}{PuncAttack}  & 0.1482              & 161.9\%             & \multicolumn{1}{c|}{29.8}          & 0.56727            & 0.864           & 0.692              & 853.9                    & 1.9                          \\
\multicolumn{1}{c|}{}                        & \multicolumn{1}{c|}{BertAttack}  & 0.1643              & 190.3\%             & \multicolumn{1}{c|}{70.4}          & 0.56620            & 0.852           & 0.712              & 2036.6                   & \textbf{1.8}                 \\ \midrule
\multicolumn{1}{c|}{\multirow{7}{*}{Toys}}   & \multicolumn{1}{c|}{Clean}       & 0.5548              & -                   & \multicolumn{1}{c|}{-}             & 0.56822            & 1.000           & 1.000              & 4060.4                   & -                            \\
\multicolumn{1}{c|}{}                        & \multicolumn{1}{c|}{ChatGPT}     & 0.5602              & 1.0\%               & \multicolumn{1}{c|}{-}             & 0.56666            & 0.793           & 0.454              & \textbf{1067.1}          & -                            \\
\multicolumn{1}{c|}{}                        & \multicolumn{1}{c|}{Trivial}      & 0.5245              & -5.5\%              & \multicolumn{1}{c|}{-}             & 0.56822            & \textbf{0.880}  & \textbf{0.852}     & 7874.4                   & -                            \\
\multicolumn{1}{c|}{}                        & \multicolumn{1}{c|}{Deepwordbug} & 0.6786              & 22.3\%              & \multicolumn{1}{c|}{\textbf{21.5}} & 0.56822            & 0.650           & 0.493              & 8181.2                   & 2.5                          \\
\multicolumn{1}{c|}{}                        & \multicolumn{1}{c|}{TextFooler}  & \textbf{0.7098}     & \textbf{27.9\%}     & \multicolumn{1}{c|}{45.6}          & \textbf{0.56886}   & 0.729           & 0.577              & 4953.7                   & 2.1                          \\
\multicolumn{1}{c|}{}                        & \multicolumn{1}{c|}{PuncAttack}  & 0.6624              & 19.4\%              & \multicolumn{1}{c|}{22.8}          & 0.56733            & 0.865           & 0.659              & 3418.2                   & 1.7                          \\
\multicolumn{1}{c|}{}                        & BertAttack                       & 0.6798              & 22.5\%              & 69.5                               & 0.56850            & 0.874           & 0.733              & 8047.0                   & \textbf{1.5}                 \\ \cmidrule(l){2-10} 
\end{tabular}
}
\caption{Performance comparison of attacking \textbf{TALLRec} where Rel. Impro. denotes relative improvement against clean setting. The best result is in \textbf{boldface}.}
\label{tab:tallrec}
\end{table*}

\begin{table*}[]
\resizebox*{\textwidth}{!}{
\begin{tabular}{@{}cccccccccc@{}}
\toprule
\multicolumn{1}{l}{\multirow{2}{*}{Dataset}} & \multirow{2}{*}{Method}          & \multicolumn{3}{c}{Effectiveness}                                              & \multicolumn{5}{c}{Stealthiness}                                                                                    \\ \cmidrule(l){3-5} \cmidrule(l){6-10} 
\multicolumn{1}{l}{}                         &                                  & Propensity $\uparrow$ & Rel. Impro. $\uparrow$ & \# queries $\downarrow $           & NDCG@10 $\uparrow$ & Cos. $\uparrow$ & Rouge-l $\uparrow$ & Perplexity $\downarrow $ & \# pert. words $\downarrow $ \\ \midrule
\multicolumn{1}{c|}{\multirow{7}{*}{Sports}} & \multicolumn{1}{c|}{Clean}       & 0.35137             & 0.0\%               & \multicolumn{1}{c|}{-}            & 0.28782            & 1.000           & 1.000              & 2158.7                   & -                           \\
\multicolumn{1}{c|}{}                        & \multicolumn{1}{c|}{ChatGPT}     & 0.36891             & 5.0\%               & \multicolumn{1}{c|}{-}            & 0.28777            & 0.794           & 0.499              & \textbf{1770.9}          & -                           \\
\multicolumn{1}{c|}{}                        & \multicolumn{1}{c|}{Trivial}      & 0.34813             & -0.9\%              & \multicolumn{1}{c|}{-}            & 0.28764            & \textbf{0.896}  & \textbf{0.869}     & 4376.6                   & -                           \\
\multicolumn{1}{c|}{}                        & \multicolumn{1}{c|}{Deepwordbug} & 0.41533             & 18.2\%              & \multicolumn{1}{c|}{\textbf{39.3}} & 0.28765            & 0.637           & 0.381              & 7417.3                   & 4.9                          \\
\multicolumn{1}{c|}{}                        & \multicolumn{1}{c|}{TextFooler}  & \textbf{0.42784}    & \textbf{21.8\%}     & \multicolumn{1}{c|}{91.2}          & 0.28768            & 0.688           & 0.499              & 2494.9                   & 4.1                          \\
\multicolumn{1}{c|}{}                        & \multicolumn{1}{c|}{PuncAttack}  & 0.39292             & 11.8\%              & \multicolumn{1}{c|}{46.7}          & \textbf{0.28789}   & 0.842           & 0.602              & 2572.7                   & \textbf{3.4}                 \\
\multicolumn{1}{c|}{}                        & \multicolumn{1}{c|}{BertAttack}  & 0.41893             & 19.2\%              & \multicolumn{1}{c|}{135.7}         & 0.28781            & 0.823           & 0.598              & 9421.4                   & 3.7                          \\ \midrule
\multicolumn{1}{c|}{\multirow{7}{*}{Beauty}} & \multicolumn{1}{c|}{Clean}       & 0.08218             & 0.0\%               & \multicolumn{1}{c|}{-}            & 0.28765            & 1.000           & 1.000              & 611.6                    & -                           \\
\multicolumn{1}{c|}{}                        & \multicolumn{1}{c|}{ChatGPT}     & 0.07889             & -4.0\%              & \multicolumn{1}{c|}{-}            & 0.28733            & 0.822           & 0.516              & \textbf{501.8}           & -                           \\
\multicolumn{1}{c|}{}                        & \multicolumn{1}{c|}{Trivial}      & 0.07734             & -5.9\%              & \multicolumn{1}{c|}{-}            & \textbf{0.28761}   & \textbf{0.939}  & \textbf{0.901}     & 1189.5                   & -                           \\
\multicolumn{1}{c|}{}                        & \multicolumn{1}{c|}{Deepwordbug} & 0.23193             & 182.2\%             & \multicolumn{1}{c|}{\textbf{47.3}} & 0.28708            & 0.640           & 0.370              & 4590.1                   & 4.9                          \\
\multicolumn{1}{c|}{}                        & \multicolumn{1}{c|}{TextFooler}  & 0.25010             & 204.3\%             & \multicolumn{1}{c|}{112.1}         & 0.28691            & 0.683           & 0.460              & 1200.3                   & 4.2                          \\
\multicolumn{1}{c|}{}                        & \multicolumn{1}{c|}{PuncAttack}  & 0.20653             & 151.3\%             & \multicolumn{1}{c|}{52.2}          & 0.28693            & 0.828           & 0.594              & 1132.0                   & \textbf{3.4}                 \\
\multicolumn{1}{c|}{}                        & \multicolumn{1}{c|}{BertAttack}  & \textbf{0.29618}    & \textbf{260.4\%}    & \multicolumn{1}{c|}{146.8}         & 0.28676            & 0.821           & 0.585              & 2634.5                   & 3.5                          \\ \midrule
\multicolumn{1}{c|}{\multirow{7}{*}{Toys}}   & \multicolumn{1}{c|}{Clean}       & 0.26065             & 0.0\%               & \multicolumn{1}{c|}{-}            & 0.28587            & 1.000           & 1.000              & 4060.4                   & -                           \\
\multicolumn{1}{c|}{}                        & \multicolumn{1}{c|}{ChatGPT}     & 0.28115             & 7.9\%               & \multicolumn{1}{c|}{-}            & 0.28610            & 0.793           & 0.454              & \textbf{1967.1}          & -                           \\
\multicolumn{1}{c|}{}                        & \multicolumn{1}{c|}{Trivial}      & 0.26913             & 3.3\%               & \multicolumn{1}{c|}{-}            & 0.28614            & \textbf{0.880}  & \textbf{0.852}     & 7874.4                   & -                           \\
\multicolumn{1}{c|}{}                        & \multicolumn{1}{c|}{Deepwordbug} & 0.49867             & 91.3\%              & \multicolumn{1}{c|}{\textbf{28.0}} & 0.28619            & 0.642           & 0.490              & 8896.5                   & 3.5                          \\
\multicolumn{1}{c|}{}                        & \multicolumn{1}{c|}{TextFooler}  & \textbf{0.52492}    & \textbf{101.4\%}    & \multicolumn{1}{c|}{65.0}          & 0.28613            & 0.733           & 0.571              & 4413.1                   & 3.0                          \\
\multicolumn{1}{c|}{}                        & \multicolumn{1}{c|}{PuncAttack}  & 0.42457             & 62.9\%              & \multicolumn{1}{c|}{31.3}          & \textbf{0.28637}   & 0.860           & 0.666              & 4134.3                   & \textbf{2.4}                 \\
\multicolumn{1}{c|}{}                        & \multicolumn{1}{c|}{BertAttack}  & 0.46528             & 78.5\%              & \multicolumn{1}{c|}{80.0}          & 0.28597            & 0.855           & 0.690              & 9618.9                   & \textbf{2.4}                 \\ \bottomrule
\end{tabular}
}
\caption{Performance comparison of attacking \textbf{CoLLM} where Rel. Impro. denotes relative improvement against clean setting. The best result is in \textbf{boldface}}
\label{tab:collm}
\end{table*}

\begin{table*}[]
\resizebox*{\textwidth}{!}{
\begin{tabular}{@{}cccccccccc@{}}
\toprule
\multicolumn{1}{l}{\multirow{2}{*}{Dataset}} & \multirow{2}{*}{Method}          & \multicolumn{3}{c}{Effectiveness}                                              & \multicolumn{5}{c}{Stealthiness}                                                                                    \\ \cmidrule(l){3-5}  \cmidrule(l){6-10} 
\multicolumn{1}{l}{}                         &                                  & Exposure $\uparrow$ & Rel. Impro. $\uparrow$ & \# queries $\downarrow $           & NDCG@10 $\uparrow$ & Cos. $\uparrow$ & Rouge-l $\uparrow$ & Perplexity $\downarrow $ & \# pert. words $\downarrow $ \\ \midrule
\multicolumn{1}{c|}{\multirow{7}{*}{Sports}} & \multicolumn{1}{c|}{Clean}       & 0.00261             & -                   & \multicolumn{1}{c|}{-}             & 0.01252            & 1.000           & 1.000              & 2158.7                   & -                            \\
\multicolumn{1}{c|}{}                        & \multicolumn{1}{c|}{ChatGPT}     & 0.00263             & 0.8\%               & \multicolumn{1}{c|}{-}             & \textbf{0.01237}   & 0.794           & 0.499              & \textbf{1770.9}          & -                            \\
\multicolumn{1}{c|}{}                        & \multicolumn{1}{c|}{Trivial}      & 0.00219             & -16.1\%             & \multicolumn{1}{c|}{-}             & \textbf{0.01237}   & \textbf{0.896}  & \textbf{0.869}     & 4376.6                   & -                            \\
\multicolumn{1}{c|}{}                        & \multicolumn{1}{c|}{Deepwordbug} & 0.00835             & 220.0\%             & \multicolumn{1}{c|}{\textbf{40.1}} & 0.01232            & 0.778           & 0.608              & 5086.0                   & 3.1                          \\
\multicolumn{1}{c|}{}                        & \multicolumn{1}{c|}{TextFooler}  & 0.01074             & 311.8\%             & \multicolumn{1}{c|}{94.0}          & 0.01228            & 0.775           & 0.595              & 2030.6                   & 3.2                          \\
\multicolumn{1}{c|}{}                        & \multicolumn{1}{c|}{PuncAttack}  & 0.00932             & 257.2\%             & \multicolumn{1}{c|}{60.0}          & 0.01235            & 0.864           & 0.672              & 2747.2                   & \textbf{2.6}                 \\
\multicolumn{1}{c|}{}                        & \multicolumn{1}{c|}{BertAttack}  & \textbf{0.01111}    & \textbf{325.9\%}    & \multicolumn{1}{c|}{165.0}         & 0.01228            & 0.827           & 0.645              & 7382.3                   & 3.2                          \\ \midrule
\multicolumn{1}{c|}{\multirow{7}{*}{Beauty}} & \multicolumn{1}{c|}{Clean}       & 0.00432             & -                   & \multicolumn{1}{c|}{-}             & 0.03022            & 1.000           & 1.000              & 611.6                    & -                            \\
\multicolumn{1}{c|}{}                        & \multicolumn{1}{c|}{ChatGPT}     & 0.00356             & -17.7\%             & \multicolumn{1}{c|}{-}             & 0.02915            & 0.822           & 0.516              & \textbf{501.8}           & -                            \\
\multicolumn{1}{c|}{}                        & \multicolumn{1}{c|}{Trivial}      & 0.00390             & -9.7\%              & \multicolumn{1}{c|}{-}             & \textbf{0.03022}   & \textbf{0.939}  & \textbf{0.901}     & 1189.5                   & -                            \\
\multicolumn{1}{c|}{}                        & \multicolumn{1}{c|}{Deepwordbug} & 0.01348             & 212.0\%             & \multicolumn{1}{c|}{\textbf{49.4}} & 0.02960            & 0.744           & 0.591              & 3626.9                   & 4.1                          \\
\multicolumn{1}{c|}{}                        & \multicolumn{1}{c|}{TextFooler}  & 0.01886             & 336.6\%             & \multicolumn{1}{c|}{116.3}         & 0.02926            & 0.758           & 0.567              & 1372.3                   & 4.5                          \\
\multicolumn{1}{c|}{}                        & \multicolumn{1}{c|}{PuncAttack}  & 0.01270             & 194.0\%             & \multicolumn{1}{c|}{72.9}          & 0.02959            & 0.853           & 0.679              & 1115.9                   & \textbf{3.3}                 \\
\multicolumn{1}{c|}{}                        & \multicolumn{1}{c|}{BertAttack}  & \textbf{0.01949}    & \textbf{351.0\%}    & \multicolumn{1}{c|}{214.8}         & 0.02928            & 0.822           & 0.642              & 2288.2                   & 4.0                          \\ \midrule
\multicolumn{1}{c|}{\multirow{7}{*}{Toys}}   & \multicolumn{1}{c|}{Clean}       & 0.00429             & -                   & \multicolumn{1}{c|}{-}             & 0.03626            & 1.000           & 1.000              & 4060.4                   & \textbf{-}                   \\
\multicolumn{1}{c|}{}                        & \multicolumn{1}{c|}{ChatGPT}     & 0.00392             & -8.6\%              & \multicolumn{1}{c|}{-}             & 0.03580            & 0.793           & 0.454              & \textbf{1967.1}          & -                            \\
\multicolumn{1}{c|}{}                        & \multicolumn{1}{c|}{Trivial}      & 0.00407             & -5.1\%              & \multicolumn{1}{c|}{-}             & \textbf{0.03623}   & \textbf{0.880}  & \textbf{0.852}     & 7874.4                   & -                            \\
\multicolumn{1}{c|}{}                        & \multicolumn{1}{c|}{Deepwordbug} & 0.01268             & 195.7\%             & \multicolumn{1}{c|}{\textbf{33.6}} & 0.03610            & 0.703           & 0.542              & 11045.3                  & 3.1                          \\
\multicolumn{1}{c|}{}                        & \multicolumn{1}{c|}{TextFooler}  & \textbf{0.01725}    & \textbf{302.5\%}    & \multicolumn{1}{c|}{86.8}          & 0.03596            & 0.709           & 0.525              & 4762.2                   & 3.3                          \\
\multicolumn{1}{c|}{}                        & \multicolumn{1}{c|}{PuncAttack}  & 0.01214             & 183.1\%             & \multicolumn{1}{c|}{41.1}          & 0.03577            & 0.845           & 0.653              & 4350.4                   & \textbf{2.4}                 \\
\multicolumn{1}{c|}{}                        & \multicolumn{1}{c|}{BertAttack}  & 0.01381             & 222.1\%             & \multicolumn{1}{c|}{116.6}         & 0.03599            & 0.829           & 0.661              & 12640.6                  & 2.6                          \\ \bottomrule
\end{tabular}
}
\caption{Performance comparison of attacking \textbf{fintuned Recformer} where Rel. Impro. denotes relative improvement against clean setting. The best result is in \textbf{boldface}.}
\label{tab:recformer-ft}
\end{table*}

\begin{table*}[]
\centering
\small

\begin{tabular}{@{}ccccccc@{}}
\toprule
           & \multicolumn{2}{c}{Sports}          & \multicolumn{2}{c}{Beauty}          & \multicolumn{2}{c}{Toys}            \\ \cmidrule(r){2-3} \cmidrule(r){4-5} \cmidrule(r){6-7}
Attack     & AUC              & Propensity       & AUC              & Propensity       & AUC              & Propensity       \\ \midrule
Clean      & 0.58489          & 0.03994          & 0.56758          & 0.05659          & 0.56822          & 0.55476          \\
Random     & 0.53550          & 0.05450          & \textbf{0.56603} & 0.08472          & 0.56394          & 0.36513          \\
Bandwagon  & 0.55781          & 0.04082          & 0.55167          & 0.00757          & 0.55189          & 0.54899          \\
Aush       & 0.57643          & 0.02868          & 0.55416          & 0.00265          & 0.55416          & 0.52596          \\
LegUP      & 0.54123          & 0.04361          & 0.53604          & 0.06628          & 0.55946          & 0.57901          \\
TextFooler & \textbf{0.58395} & \textbf{0.10926} & 0.56581          & \textbf{0.17238} & \textbf{0.56886} & \textbf{0.70980} \\ \bottomrule
\end{tabular}
\caption{Shilling attacks on \textbf{TALLRec}.}
\label{tab:shilling-tallrec}
\end{table*}

\begin{table*}[]
\centering
\small

\begin{tabular}{@{}ccccccc@{}}
\toprule
           & \multicolumn{2}{c}{Sports}          & \multicolumn{2}{c}{Beauty}          & \multicolumn{2}{c}{Toys}            \\ \cmidrule(r){2-3} \cmidrule(r){4-5} \cmidrule(r){6-7}
Attack     & AUC              & Propensity       & AUC              & Propensity       & AUC              & Propensity       \\ \midrule
Clean      & 0.58331          & 0.57760          & 0.56256          & 0.22535          & 0.58622          & 0.38361          \\
Random     & 0.55275          & 0.57192          & 0.52894          & 0.27450          & 0.54497          & 0.24871          \\
Bandwagon  & 0.57669          & 0.56773          & 0.53884          & 0.23739          & 0.55721          & 0.34892          \\
Aush       & 0.57767          & 0.55084          & 0.54857          & 0.24656          & 0.55373          & 0.49610          \\
LegUP      & 0.57648          & 0.54010          & 0.53802          & 0.19230          & 0.54504          & 0.39779          \\
TextFooler & \textbf{0.58285} & \textbf{0.65581} & \textbf{0.55993} & \textbf{0.38677} & \textbf{0.58706} & \textbf{0.57592} \\ \bottomrule
\end{tabular}
\caption{Shilling attacks on \textbf{CoLLM}.}
\label{tab:shilling-collm}
\end{table*}

\section{Case Studies}
In this section, we present examples of attacks and defenses against RecFormer as a victim model, as shown in Table \ref{tab:case1} - \ref{tab:case-1}.

\begin{table*}
\centering
\small
\begin{tblr}{
  cells = {c},
  hline{1,13} = {-}{0.08em},
  hline{2, 3, 5, 7, 9, 11} = {-}{},
}
Model         & Text                                                     & Exposure \\
Clean         & Little People Surprise Sounds Fun Park                   & 0.0111   \\
Trivial        & Little People Surprise Sounds Fun Park \textcolor{red}{good fantastic} & 0.0151   \\
GPT & \textcolor{red}{Exciting} Sounds Fun Park for Little \textcolor{red}{Ones}                 & 0.0201   \\
DeepwordBug   & ~Little \textcolor{red}{ePople} Surprise Sounds Fun Park                  & 0.0701   \\
+Defense   & ~Little People Surprise Sounds Fun Park                  & 0.0111   \\

PunAttack     & Little \textcolor{red}{P-eople} Surprise Sounds Fun Park                  & 0.0496   \\
+Defense   & ~Little People Surprise Sounds Fun Park                  & 0.0111   \\

Textfooler    & Little \textcolor{red}{Inhabitants} Surprise Sounds Fun Park            & 0.0702   \\
+Defense   & ~Surprising \textcolor{red}{Audible Comic} Park: Little \textcolor{red}{Inhabitants}                  & 0.0586   \\

BertAttack    & Little \textcolor{red}{joe} Surprise Sounds Fun Park                      & 0.0326  \\
+Defense   & ~Little \textcolor{red}{Joe's} Surprise \textcolor{red}{Sound} Fun Park                  & 0.0198   \\
\end{tblr}
\caption{Item “B00008PVZG” in the Amazon-Toys dataset. The red part points out the differences from the original text.}
\label{tab:case1}
\end{table*}

\begin{table*}
\centering
\small
\begin{tblr}{
  cells = {c},
  hline{1,13} = {-}{0.08em},
  hline{2, 3, 5, 7, 9, 11} = {-}{},
}
Model         & Text                                                & Exposure \\
Clean         & Fisher-Price Fun-2-Learn Smart Tablet             & 0.0076   \\
Trivial        & Fisher-Price
  Fun-2-Learn Smart Tablet \textcolor{red}{better selling} & 0.0095   \\
GPT & \textcolor{red}{Interactive}
  Learning Tablet \textcolor{red}{for Kids}              & 0.0335   \\
DeepwordBug   & Fisher-Price Fun-2-Learn \textcolor{red}{Smar Tmblet}              & 0.0335   \\
+Defense   & Fisher-Price Fun-2-Learn Smart Tablet              & 0.0076   \\

PunAttack     & Fisher-Price Fun--2-Learn \textcolor{red}{Sm'art} Tablet           & 0.0285   \\
+Defense     & Fisher-Price Fun-2-Learn Smart Tablet              & 0.0076   \\

Textfooler    & Fisher-Price  Fun-2-Learn \textcolor{red}{Canny Table}              & 0.0768   \\
+Defense    & Fisher-Price Fun-2-Learn \textcolor{red}{Canine Table}              & 0.0756   \\

BertAttack    & Fisher-Price Fun-2-Learn \textcolor{red}{this} Tablet              & 0.0262   \\
+Defense    & Fisher-Price \textcolor{red}{Fun-2-Learn Tablet}              & 0.0190   

\end{tblr}
\caption{Item “B005XVCTAU” in the Amazon-Toys dataset. The red part points out the differences from the original text.}
\end{table*}

\begin{table*}
\centering
\small
\begin{tblr}{
  cells = {c},
  hline{1,13} = {-}{0.08em},
  hline{2, 3, 5, 7, 9, 11} = {-}{},
}
Model         & Text                                                                               & Exposure \\
Clean         & Salon
  Grafix Healthy Hair Nutrition Cleansing Conditioner, 12 oz                 & 0.0321   \\
Trivial        & Salon
  Grafix Healthy Hair Nutrition Cleansing Conditioner, 12 oz \textcolor{red}{fantastic loved} & 0.0103   \\
GPT & \textcolor{red}{Nourishing Hair Care:} Salon Grafix Cleansing Conditioner, 12 oz                  & 0.0331   \\
DeepwordBug   & Salon Grafix Healthy Hair \textcolor{red}{Nutirtion} Cleansing Conditioner, 12 oz                 & 0.1020   \\
+Defense   & Salon Grafix Healthy Hair Nutrition Cleansing Conditioner, 12 oz, 12 oz                 & 0.0321   \\

PunAttack     & Salon Grafix Healthy Hair \textcolor{red}{Nutrit-ion} Cleansing Conditioner, 12 oz                       & 0.0866   \\
+Defense   & Salon Grafix Healthy Hair Nutrition Cleansing Conditioner, 12 oz                 & 0.0321   \\

Textfooler    & Salon Grafix Healthy Hair \textcolor{red}{Nourishment} Cleansing Conditioner, 12 oz                        & 0.1438   \\
+Defense   & Salon Grafix Healthy Hair \textcolor{red}{Nourishing} Cleansing Conditioner, 12 oz                 & 0.1135   \\

BertAttack    & Salon Grafix Healthy Hair \textcolor{red}{style} Cleansing Conditioner, 12 oz                            & 0.1172   \\
+Defense   & Salon Grafix Healthy Hair \textcolor{red}{Style} Cleansing Conditioner, 12 oz.                 & 0.1139   \\

\end{tblr}
\caption{Item “B007MNYY14” in the Amazon-Beauty dataset. The red part points out the differences from the original text.}
\end{table*}

\begin{table*}
\centering
\small
\begin{tblr}{
  cells = {c},
  hline{1,13} = {-}{0.08em},
  hline{2, 3, 5, 7, 9, 11} = {-}{},
}
Model       & Text                                                                                     & Exposure \\
Clean       & Phyto Organics Set Theratin Shampoo amp; Humectin Conditioner 1L Each                  & 0.0014   \\
Trivial      & Phyto
  Organics Set Theratin Shampoo amp; Humectin Conditioner 1L Each \textcolor{red}{purchase
  cheap} & 0.0015   \\
GPT         & \textcolor{red}{Luxurious}
  Phyto Organics Set: Theratin Shampoo  Humectin Conditioner - 1L Each\textcolor{red}{!}        & 0.0019   \\
DeepwordBug & Phyto Organics \textcolor{red}{et heratin} Shampoo amp; Humectin Conditioner 1L Each                    & 0.0370   \\
+Defense & Phyto Organics \textcolor{red}{and Keratin} Shampoo \textcolor{red}{\&} Humectin Conditioner 1L Each                   & 0.0170   \\

PunAttack     & Phyto \textcolor{red}{Organic's Se't} Theratin Shampoo amp; Humectin Conditioner 1L Each               & 0.0386  \\
+Defense & Phyto Organics Set Theratin Shampoo amp; Humectin Conditioner 1L Each                    & 0.0014   \\

Textfooler   & Phyto Organic \textcolor{red}{Setting} Theratin Shampoo amp; Humectin Conditioner 1L Each               & 0.0200  \\
+Defense & Phyto Organic \textcolor{red}{Setting} Theratin Shampoo \textcolor{red}{\&} Humectin Conditioner - 1L Each                   & 0.0272   \\

BertAttack  & Phyto Organics \textcolor{red}{for} Theratin Shampoo \textcolor{red}{makeup}; Humectin Conditioner 1L Each              & 0.0483  \\
+Defense & Phyto Organics \textcolor{red}{for} Theratin Shampoo \textcolor{red}{\& Hair} Conditioner \textcolor{red}{-} 1L Each                    & 0.0540   

\end{tblr}
\caption{Item “B0030UG27W” in the Amazon-Beauty dataset. The red part points out the differences from the original text.}
\end{table*}

\begin{table*}
\centering
\small
\begin{tblr}{
  cells = {c},
  hline{1,13} = {-}{0.08em},
  hline{2, 3, 5, 7, 9, 11} = {-}{},
}
Model       & Text                                                         & Exposure \\
Clean       & GAIAM
  Toeless Grippy Yoga Socks Toesocks                   & 0.0017   \\
Trivial      & GAIAM
  Toeless Grippy Yoga Socks Toesocks \textcolor{red}{wonderful product} & 0.0032   \\
GPT         & GAIAM
  Grippy Toeless Yoga Socks - \textcolor{red}{Ultimate} Toesocks        & 0.0036   \\
DeepwordBug & GAIAM Toeless Grippy Yoga \textcolor{red}{oScks} Toesocks                   & 0.0071   \\
+Defense & GAIAM Toeless GripBpy Yoga Socks - \textcolor{red}{Quality Socks}                   & 0.0041   \\

PunAttack   & \textcolor{red}{GA-IAM} Toeless \textcolor{red}{Grip-py} Yoga Sock's Toesocks               & 0.0085   \\
+Defense & \textcolor{red}{GA-IAM} Toeless Grippy Yoga Sock's Toesocks                   & 0.0053   \\

TextFooler  & GAIAM Toeless Grippy Yoga \textcolor{red}{Sock} Toesocks                    & 0.0111   \\
+Defense & \textcolor{red}{Gaiam} Toeless Grippy Yoga Socks \textcolor{red}{- Toe Socks}           & 0.0031   \\

BertAttack  & GAIAM Toeless Grippy Yoga \textcolor{red}{with} Toesocks                    & 0.0080   \\
+Defense & GAIAM Toeless \textcolor{red}{Grip} Yoga Socks with Toesocks                   & 0.0056   
\end{tblr}
\caption{Item “B008EADJPG” in the Amazon-Sports dataset. The red part points out the differences from the original text.}
\end{table*}

\begin{table*}
\centering
\small
\begin{tblr}{
  cells = {c},
  hline{1,13} = {-}{0.08em},
  hline{2, 3, 5, 7, 9, 11} = {-}{},
}
Model       & Text                                                                                           & Exposure \\
Clean       & BladesUSA E419-PP Polypropylene Karambit Training Knife 6.7-Inch Overall                     & 0.0014   \\
Trivial      & BladesUSA E419-PP Polypropylene Karambit Training Knife 6.7-Inch Overall \textcolor{red}{quality excellent} & 0.0012   \\
GPT         & \textcolor{red}{Ultimate} Training Knife: BladesUSA E419-PP Karambit - \textcolor{red}{Unbeatable Performance!}             & 0.0041   \\
DeepwordBug & BladesUSA E419-PP \textcolor{red}{Polyproylene} Karambit Training Knife 6.7-Inch \textcolor{red}{Ovearll}                     & 0.0158   \\
+ Defense & BladesUSA E419-PP Polypropylene Karambit Training Knife 6.7-Inch Overall                     & 0.0014   \\
PunAttack   & \textcolor{red}{Bla'desUSA} E419-PP \textcolor{red}{Polypropy'lene} Karambit Training Knife 6.7-Inch Overall                 & 0.0117   \\
  + Defense & BladesUSA E419-PP Polypropylene Karambit Training Knife 6.7-Inch Overall                     & 0.0014   \\
TextFooler  & BladesUSA E419-PP Polypropylene Karambit Training \textcolor{red}{Knifes} 6.7-Inch Overall                    & 0.0036   \\
+ Defense  & BladesUSA E419-PP Polypropylene Karambit Training \textcolor{red}{Knifes} 6.7-Inch Overall                                 & 0.0036 \\
BertAttack  & BladesUSA E419-PP \textcolor{red}{a} Karambit Training Knife 6.7-Inch Overall                                 & 0.0095   \\
  + Defense & BladesUSA E419-PP\textcolor{red}{: A} Karambit Training Knife with 6.7-Inch Overall Length & 0.0082   \\
\end{tblr}
\caption{Item “B0089AH12I” in the Amazon-Sports dataset. The red part points out the differences from the original text.}
\label{tab:case-1}
\end{table*}

\end{document}